%% file: main.tex
\documentclass{article}

\usepackage[preprint]{icml2026/icml2026}
\makeatletter
\renewcommand{\ICML@preprint}{}
\makeatother

\input{preamble}

\icmltitlerunning{Learning from Saturated Data: Signals Beyond Correctness for LLM Training}

\begin{document}

\twocolumn[
  \icmltitle{Learning from Saturated Data: Signals Beyond Correctness for LLM Training}

  \begin{icmlauthorlist}
    \icmlauthor{Hanno Hiss}{eth}
    \icmlauthor{Jasper Dekoninck}{eth}
    \icmlauthor{Martin Vechev}{eth}
  \end{icmlauthorlist}

  \icmlaffiliation{eth}{ETH Zurich}

  \icmlcorrespondingauthor{Hanno Hiss}{hhiss@ethz.ch}

  \icmlkeywords{Machine Learning, ICML}

  \vskip 0.3in
]

\printAffiliationsAndNotice{Code: \url{https://github.com/eth-sri/saturated-learning}. }

\input{sections/abstract}
\input{sections/introduction}
\input{sections/related_work}
\input{sections/methods}
\input{sections/results}
\input{sections/discussion}
\input{sections/conclusion}
\input{sections/impact_statement}
\input{sections/acknowledgements}
\bibliography{references}
\bibliographystyle{icml2026/icml2026}
\newpage
\input{sections/appendix}

\end{document}

%% file: preamble.tex
\usepackage{colortbl}
\usepackage{tabularx}
\usepackage{microtype}
\usepackage{graphicx}
\usepackage{pdfpages}
\usepackage{subcaption} 
\usepackage{enumitem}
\usepackage{placeins}

\usepackage{float}            
\usepackage{tcolorbox}
\tcbuselibrary{listings,breakable,skins}  
\newtcblisting{prompt}[1]{
  title={#1},
  breakable,
  skin=enhanced jigsaw,
  colback=white,
  colframe=teal!60!black,
  coltitle=white,
  fonttitle=\bfseries,
  listing only,
  listing options={
    basicstyle=\small\ttfamily,
    breaklines=true,
    columns=flexible,
  },
  left=6pt, right=6pt, top=4pt, bottom=4pt,
  extras first={sharp corners=south},
  extras middle={sharp corners},
  extras last={sharp corners=north},
}

\newcounter{takeaway}
\newtcolorbox{takeaway}[2][]{
  before upper={},
  breakable,
  enhanced,
  colback=gray!5,
  colframe=black!80,
  arc=4pt,
  boxrule=1.5pt,
  left=8pt, right=8pt, top=12pt, bottom=5pt,
  attach boxed title to top left={yshift=-\tcboxedtitleheight/2, xshift=6pt},
  boxed title style={
    colback=black!80,
    colframe=black!80,
    arc=3pt,
    boxrule=0pt,
    left=6pt, right=6pt, top=2pt, bottom=2pt,
  },
  fonttitle=\bfseries\sffamily\small\color{white},
  title={\refstepcounter{takeaway}Takeaway \thetakeaway},
  #1,
}

\newtcolorbox{taskexample}[1]{
  title={#1},
  breakable,
  skin=enhanced jigsaw,
  colback=teal!5,
  colframe=teal!60!black,
  coltitle=white,
  fonttitle=\bfseries,
  arc=2pt,
  left=8pt, right=8pt, top=6pt, bottom=6pt,
  fontupper=\small,
}
\newtcolorbox{taskexampletwo}[1]{
  title={#1},
  skin=enhanced jigsaw,
  colback=teal!5,
  colframe=teal!60!black,
  coltitle=white,
  fonttitle=\bfseries,
  arc=2pt,
  left=8pt, right=8pt, top=6pt, bottom=6pt,
  fontupper=\small,
}

\usepackage{tikz}
\usetikzlibrary{arrows.meta, positioning, shapes.geometric, calc, shapes.callouts, decorations.pathreplacing}
\newcommand{\tikzfigname}[1]{}  

\usepackage{pgfplots}
\pgfplotsset{compat=1.18}
\usepgfplotslibrary{colorbrewer}
\usepackage{pgfplotstable}

\usepackage{amsmath}
\usepackage{amssymb}
\usepackage{mathtools}
\usepackage{amsthm}

\theoremstyle{definition}

\theoremstyle{remark}

\usepackage{hyperref}
\definecolor{mydarkblue}{rgb}{0,0.08,0.45}
\hypersetup{
  pdfborder=0 0 0,
  colorlinks=true,
  linkcolor=mydarkblue,
  citecolor=mydarkblue,
  filecolor=mydarkblue,
  urlcolor=mydarkblue,
}
\usepackage[capitalize,noabbrev]{cleveref}
\usepackage{booktabs}         

\definecolor{plotblue}{HTML}{4C72B0}
\definecolor{plotorange}{HTML}{DD8452}
\definecolor{plotgreen}{HTML}{55A868}
\definecolor{plotred}{HTML}{C44E52}
\definecolor{plotpurple}{HTML}{8172B2}
\definecolor{plotteal}{HTML}{64B5CD}

\definecolor{cellblue}{HTML}{C8E6F5}    
\definecolor{cellorange}{HTML}{FFE0B2}  

\newcommand{\heatcell}[3]{%
  \pgfmathsetmacro{\cpct}{int((#1 - #2) / (#3 - #2) * 60)}%
  \edef\tmp{\noexpand\cellcolor{teal!\cpct!white}}\tmp #1\%%
}

\newif\ifverboseprompts
\verbosepromptstrue

\newcommand{\Qeasy}{\mathcal{D}_{\mathrm{saturated}}}
\newcommand{\Qhard}{\mathcal{D}_{\mathrm{hard}}}
\newcommand{\Qstrict}{\mathcal{D}'_{\mathrm{strict}}}

\newcommand{\usedmodel}{\texttt{Qwen3-1.7B-Base}}
\newcommand{\strongmodel}{\texttt{Qwen3-30B}}

\newcommand{\note}[1]{}

\crefname{section}{Sec.}{Sec.}
\crefname{subsection}{Sec.}{Sec.}
\crefname{subsubsection}{Sec.}{Sec.}
\crefname{paragraph}{Sec.}{Sec.}
\crefname{appendix}{App.}{App.}
\crefname{equation}{Eq.}{Eq.}

%% file: sections/abstract.tex
\begin{abstract}
The growing capabilities of large language models (LLMs) have led to the saturation of many benchmarks and training datasets used to improve them. Motivated by this, we investigate whether questions solved with perfect empirical accuracy can nevertheless be used to improve downstream performance. To do so, we replace binary correctness with two sources of more fine-grained quality signals: (1) pairwise LLM self-judgments, in which the model evaluates the relative quality of its own solutions, and (2) token-level entropy, where token-level uncertainty is used as a proxy for solution quality. We incorporate these signals into several training algorithms and evaluate them on Qwen3-1.7B-Base. When training exclusively on a simple arithmetic task, quality-based signals improve performance by up to $18.6\%$ over the base model, substantially outperforming SFT. On GSM8K, however, gains are more modest and depend strongly on the quality signal. For instance, self-judgments show poor agreement with a stronger external judge and can even degrade performance below the base model. Overall, our results suggest that quality-based training can extract useful signal from saturated questions for base models, but that applying such signals to more complex tasks requires careful calibration and further study.
\end{abstract}

%% file: sections/introduction.tex
\section{Introduction}
\label{sec:introduction}

\input{figures/accept_figure}

Large language models (LLMs) have advanced rapidly, with successive generations achieving substantial gains on mathematical reasoning, code synthesis, and other challenging benchmarks~\citep{brownLanguageModelsAre2020,deepseek-aiDeepSeekR1IncentivizingReasoning2025,yangQwen3TechnicalReport2025}. Yet the very success of these models creates a new problem: as models grow more capable, they saturate the benchmarks and training sets used to improve them. As a result, training data that the model already solves reliably are not useful in current training regimes.

Recent work has begun to acknowledge this problem. \citet{wangReinforcementLearningReasoning2025} show that reinforcement learning (RL) with a single training example can improve reasoning performance beyond training-set saturation, though most of the reported gains are attributed to format correction rather than actual reasoning improvements.
Further, \citet{agarwalUnreasonableEffectivenessEntropy2025} demonstrate that entropy minimization alone can match GRPO-level gains without any labeled data, but they use entropy as a global training objective rather than as a quality metric to differentiate among correct completions. This leads us to ask:

\begin{center}
  \begin{minipage}{0.95\linewidth}
    \centering
    \textbf{Can we extract a learning signal from questions the model empirically solves with perfect accuracy?}
  \end{minipage}
\end{center}

As shown in \cref{fig:accept-figure}, we study this question by examining quality metrics that distinguish among correct solutions, since even correct model responses can differ in reasoning clarity, conciseness, and computational efficiency.
We leverage these quality differences by using them in training and investigate whether the accuracy of the model on harder questions can be improved by training on quality-ranked solutions to saturated questions.

On these saturated questions, we rank correct completions by quality using two complementary mechanisms.
First, we use a \textit{self-judge} that scores completions pairwise by prompting the model to select the better of two completions. Second, we use \textit{self-uncertainty} as a proxy for quality, ranking completions by the inverse mean token-level entropy of the model's output distribution. We then use these signals with two different training algorithms: Direct Preference Optimization (DPO)~\citep{rafailovDirectPreferenceOptimization2024}, which trains on chosen-rejected pairs constructed from the rankings, and Logistic-Weighted RRHF ($\sigma$-RRHF), a variant of RRHF~\citep{yuanRRHFRankResponses2023} that uses a logistic weighting function to assign higher importance to higher-ranked completions while still learning from all samples.

In our experiments, we train on the \texttt{chain sum} arithmetic task from ReasoningGym~\citep{stojanovskiREASONINGGYMReasoning2025} and on GSM8K~\citep{cobbeTrainingVerifiersSolve2021}, with \usedmodel{}~\citep{yangQwen3TechnicalReport2025} as the base model.
Our main findings are:
\begin{enumerate}
  \item \textbf{Easy-to-hard transfer on \texttt{chain sum}:} $\sigma$-RRHF with inverse-entropy-ranked completions improves $\text{pass}@1$ by $+18.6\%$ over the base model when trained on the saturated subset of the evaluation distribution, and still yields $+11.7\%$ over base when trained on saturated samples from a strictly easier distribution.
  \item \textbf{Scorer failures on GSM8K:} In contrast, the self-judge anti-correlates with a strong external judge on GSM8K and degrades $\sigma$-RRHF below the base model. Inverse entropy helps the base model (\(+3.3\%\)), but fails to do so for its instruction-tuned variant.
\end{enumerate}

Together, the results show that quality differences among correct completions can extend training past saturation, with scorer reliability as the central bottleneck. Our key contributions are:
\begin{itemize}
  \item We identify saturated questions as an important training setting and identify quality-based signals as a promising way to train on these questions.
  \item We compare self-judging and inverse-entropy rankings across synthetic and GSM8K tasks, showing how scorer reliability determines whether easy-question training helps or hurts.
\end{itemize}

%% file: figures/accept_figure.tex

\begin{figure*}[t]
  \centering
  \tikzfigname{accept_figure}
  \input{figures/accept_figure_inner.tex}
  \caption{%
    Quality signals on saturated questions. \textbf{(1)}  \emph{Correctness}: when all
    sampled completions are correct, binary reward provides no contrast. 
    \textbf{(2)} \emph{LLM-as-a-Judge}: a judge runs all pairwise comparisons, the win count $w_i$ ranks
    completions by judged quality.
    \textbf{(3)} \emph{Inverse Entropy}: mean token-level entropy along each completion gives a judge-free quality proxy.
    }
  \label{fig:accept-figure}
\end{figure*}
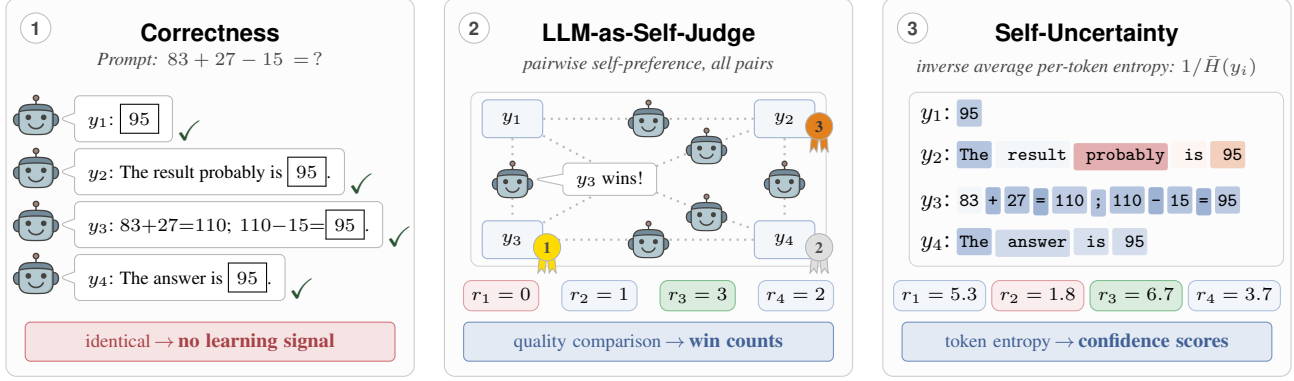

%% file: figures/accept_figure_inner.tex

\def\chkc{\textcolor{plotgreen!50!black}{\scalebox{1.5}{$\checkmark$}}}
\def\acScoreY{-3.94}
\def\acColA{-1.95}
\def\acColB{-0.65}
\def\acColC{0.65}
\def\acColD{1.95}
\def\acLabelX{-2.3}
\def\acTokenY{-1.60}
\def\acYlabX{-1.80}

\providecommand{\medalribbon}[3]{%
  \fill[#2, draw=black!30, line width=0.1pt]
  ([xshift=-1.0mm,yshift=-1.0mm]#1)
  -- ([yshift=-1.0mm]#1)
  -- ([xshift=-0.3mm,yshift=-3.4mm]#1)
  -- ([xshift=-0.7mm,yshift=-2.8mm]#1)
  -- ([xshift=-1.2mm,yshift=-3.4mm]#1)
  -- cycle;
  \fill[#3, draw=black!30, line width=0.1pt]
  ([yshift=-1.0mm]#1)
  -- ([xshift=1.0mm,yshift=-1.0mm]#1)
  -- ([xshift=1.2mm,yshift=-3.4mm]#1)
  -- ([xshift=0.7mm,yshift=-2.8mm]#1)
  -- ([xshift=0.3mm,yshift=-3.4mm]#1)
  -- cycle;
}

\begin{tikzpicture}[
    >=Latex,
    panel/.style={
      draw=gray!40, rounded corners=4pt, fill=gray!2,
      minimum width=5.4cm, minimum height=5.0cm,
      anchor=north west,
    },
    pnum/.style={
      circle, draw=gray!50, fill=white, inner sep=0pt,
      minimum size=4mm, font=\scriptsize\bfseries\sffamily,
      text=gray!55!black,
    },
    ptitle/.style={font=\bfseries\sffamily\small, align=center, inner ysep=1pt},
    psub/.style={font=\scriptsize\itshape, text=gray!55!black, align=center, inner ysep=1pt},
    yb/.style={
      draw=plotblue!55, rounded corners=2pt, fill=plotblue!6,
      inner xsep=2pt, inner ysep=1.5pt, font=\scriptsize,
      minimum width=8mm, minimum height=5mm, anchor=center,
      align=center,
    },
    ytext/.style={font=\scriptsize, anchor=west, inner xsep=0pt, inner ysep=0pt},
    sbox/.style={
      draw=plotblue!45, rounded corners=3pt, fill=plotblue!7,
      inner xsep=3pt, inner ysep=2pt,
      font=\scriptsize, anchor=center,
      minimum width=9mm, minimum height=4mm, align=center,
    },
    sbest/.style={sbox, draw=plotgreen!75, fill=plotgreen!22,
    font=\scriptsize\bfseries},
    sworst/.style={sbox, draw=plotred!55, fill=plotred!10},
    scorrect/.style={sbox, draw=plotgreen!55, fill=plotgreen!10,
    text=plotgreen!45!black},
    callout/.style={
      draw=plotred!80, rounded corners=2pt, fill=plotred!15,
      inner xsep=7pt, inner ysep=4pt, anchor=south,
      font=\scriptsize, align=center, text=plotred!85!black,
      text width=4.4cm,
    },
    calloutpos/.style={
      draw=plotblue!80, rounded corners=2pt, fill=plotblue!15,
      inner xsep=7pt, inner ysep=4pt, anchor=south,
      font=\scriptsize, align=center, text=plotblue!85!black,
      text width=4.4cm,
    },
    bubble/.style={
      draw=gray!55, rounded corners=2pt, fill=white,
      rectangle callout, callout pointer width=1.5mm,
      font=\scriptsize, inner xsep=5pt, inner ysep=3pt,
      anchor=center,
    },
    speechbubble/.style={
      draw=gray!55, rounded corners=2pt, fill=white,
      rectangle callout, callout pointer width=1.5mm,      callout relative pointer={(-0.5,-0.5)},
      font=\scriptsize, inner xsep=5pt, inner ysep=3pt,
      anchor=west, align=left,
    },
    ebar/.style={draw=none, fill=plotorange!75},
    ebox/.style={draw=gray!40, fill=white, rounded corners=1pt, inner sep=0pt},
    ysub/.style={font=\tiny, text=gray!50!black, anchor=north},
    htok/.style={
      draw=none, rounded corners=1pt, inner xsep=1pt, inner ysep=1.5pt,
      font=\scriptsize\ttfamily, anchor=base west, text=black,
      minimum height=3.6mm,
    },
    heat/.code={%
      \pgfmathsetmacro{\acHv}{#1}%
      \pgfmathparse{\acHv<0.5 ? 1 : 0}%
      \ifdim\pgfmathresult pt>0.5pt
      \pgfmathtruncatemacro{\acHp}{max(0, min(70, round((0.5-\acHv)*120)))}%
      \tikzset{fill=plotblue!\acHp}%
      \else
      \pgfmathparse{\acHv<0.8 ? 1 : 0}%
      \ifdim\pgfmathresult pt>0.5pt
      \pgfmathtruncatemacro{\acHp}{max(0, min(60, round((\acHv-0.5)*150)))}%
      \tikzset{fill=plotorange!\acHp}%
      \else
      \pgfmathtruncatemacro{\acHp}{max(25, min(70, round(25 + (\acHv-0.8)*200)))}%
      \tikzset{fill=plotred!\acHp}%
      \fi
      \fi
    },
    etag/.style={
      font=\tiny, text=gray!55!black, anchor=south, inner sep=0pt,
    },
    ylab/.style={
      font=\footnotesize, anchor=base east, inner sep=0pt,
    },
    tcell/.style={
      draw=gray!45, rounded corners=1pt,
      font=\scriptsize, inner xsep=1pt, inner ysep=1pt,
      minimum width=11mm, minimum height=3.2mm, align=center,
    },
    medal/.style={
      circle, draw=black!30, line width=0.2pt, inner sep=0pt,
      minimum size=3.6mm, font=\tiny\bfseries, anchor=center,
    },
    gold/.style={medal, fill=yellow!80!orange, text=black!70},
    silver/.style={medal, fill=gray!25, text=black!70},
    bronze/.style={medal, fill=orange!55!brown, text=black!80},
  ]

  \node[panel] (P1) at (0,0) {};
  \node[pnum] at ([shift={(4mm,-4mm)}]P1.north west) {1};
  \node[ptitle, anchor=north] (P1t) at ([yshift=-3mm]P1.north) {Correctness};
  \node[psub,   anchor=north] (P1s) at ([yshift=-0.5mm]P1t.south)
  {Prompt:\ \ $83 + 27 - 15 \,=\, ?$};


  \node[anchor=center, inner sep=0pt] (LLMgen)
  at ([shift={(-2.3,-1.55)}]P1.north)
  {\includegraphics[height=5.5mm]{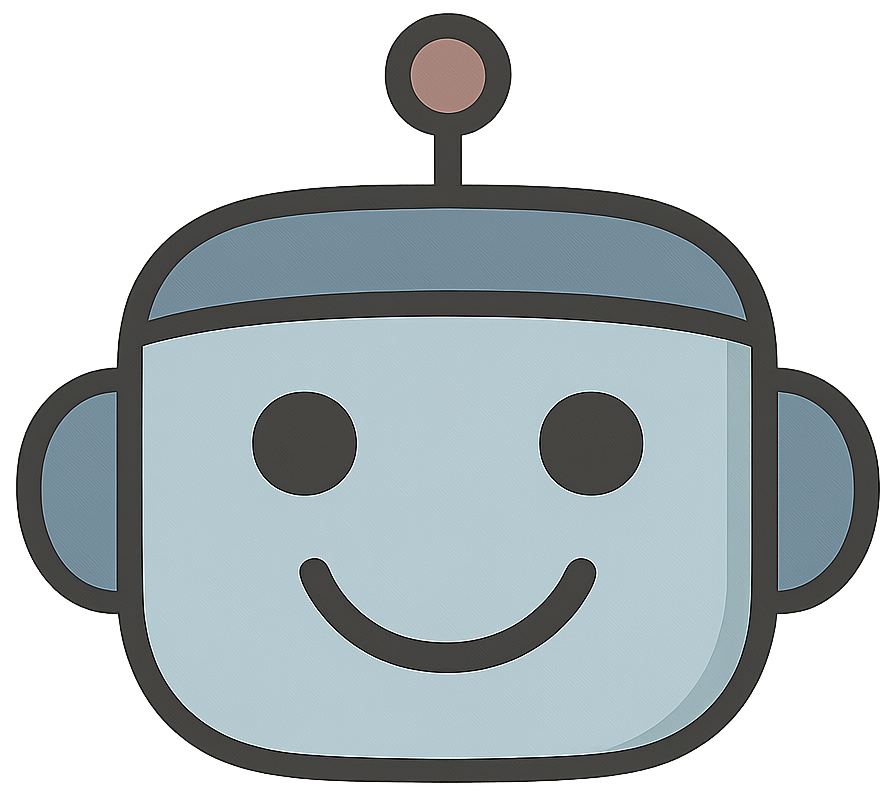}};
  \node[anchor=center, inner sep=0pt] (LLMgen)
  at ([shift={(-2.3,-3.65)}]P1.north)
  {\includegraphics[height=5.5mm]{figures/bot1.png}};
  \node[anchor=center, inner sep=0pt] (LLMgen)
  at ([shift={(-2.3,-2.25)}]P1.north)
  {\includegraphics[height=5.5mm]{figures/bot1.png}};
  \node[anchor=center, inner sep=0pt] (LLMgen)
  at ([shift={(-2.3,-2.95)}]P1.north)
  {\includegraphics[height=5.5mm]{figures/bot1.png}};

  \node[speechbubble, callout relative pointer={(-0.2, 0)}] (gen1)
  at ([shift={(-1.8,-1.60)}]P1.north) {$y_1$: $\boxed{95}$};
  \node[speechbubble, callout relative pointer={(-0.2, 0)}] (gen2)
  at ([shift={(-1.8,-3.70)}]P1.north) {$y_4$: The answer is $\boxed{95}$.};
  \node[speechbubble, callout relative pointer={(-0.2, 0)}] (gen3)
  at ([shift={(-1.8,-2.30)}]P1.north) {$y_2$: The result probably is $\boxed{95}$.};
  \node[speechbubble, callout relative pointer={(-0.2, 0)}] (gen4)
  at ([shift={(-1.8,-3.00)}]P1.north) {$y_3$: $83{+}27{=}110$;\, $110{-}15{=}\boxed{95}$.};

  \node[ytext, anchor=east] at ([shift={(0.4,-0.15)}]gen1.east) {\chkc};
  \node[ytext, anchor=east] at ([shift={(0.4,-0.15)}]gen2.east) {\chkc};
  \node[ytext, anchor=east] at ([shift={(0.4,-0.15)}]gen3.east) {\chkc};
  \node[ytext, anchor=east] at ([shift={(0.4,-0.15)}]gen4.east) {\chkc};


  \node[callout] (C1) at ([yshift=2mm]P1.south)
  {identical\,$\rightarrow$\,{\fontsize{7.5}{9}\selectfont\textbf{no learning signal}}};

  \node[panel] (P2) at (5.8,0) {};
  \node[pnum] at ([shift={(4mm,-4mm)}]P2.north west) {2};
  \node[ptitle, anchor=north] (P2t) at ([yshift=-3mm]P2.north) {LLM-as-Self-Judge};
  \node[psub,   anchor=north] (P2s) at ([yshift=-0.5mm]P2t.south)
  {pairwise self-preference, all pairs};

  \draw[gray!35, rounded corners=3pt]
  ([shift={(-2.35,-1.25)}]P2.north) rectangle ([shift={(2.35,-3.55)}]P2.north);

  \node[yb] (yb2_1) at ([shift={(-1.8,-1.60)}]P2.north) {$y_1$};
  \node[yb] (yb2_2) at ([shift={( 1.8,-1.60)}]P2.north) {$y_2$};
  \node[yb] (yb2_3) at ([shift={(-1.8,-3.20)}]P2.north) {$y_3$};
  \node[yb] (yb2_4) at ([shift={( 1.8,-3.20)}]P2.north) {$y_4$};

  \draw[-, gray!55, dotted, thick] (yb2_1) to[bend right=0] coordinate[pos=0.5] (m12) (yb2_2);
  \draw[-, gray!55, dotted, thick] (yb2_3) to[bend left=0]  coordinate[pos=0.5] (m34) (yb2_4);
  \draw[-, gray!55, dotted, thick] (yb2_1) to[bend left=0]  coordinate[pos=0.5] (m13) (yb2_3);
  \draw[-, gray!55, dotted, thick] (yb2_2) to[bend right=0] coordinate[pos=0.5] (m24) (yb2_4);
  \draw[-, gray!55, dotted, thick] (yb2_1) to[bend right=5] coordinate[pos=0.8] (m14) (yb2_4);
  \draw[-, gray!55, dotted, thick] (yb2_3) to[bend left=5]  coordinate[pos=0.8] (m23) (yb2_2);

  \coordinate (med3) at ([xshift=0.5mm, yshift=-1mm]yb2_3.east);
  \coordinate (med4) at ([xshift=0.5mm, yshift=-1mm]yb2_4.east);
  \coordinate (med2) at ([xshift=0.5mm, yshift=-1mm]yb2_2.east);
  \medalribbon{med3}{yellow!80!orange}{yellow!80!orange}
  \medalribbon{med4}{gray!25}{gray!25}
  \medalribbon{med2}{orange!55!brown}{orange!55!brown}
  \node[gold,   anchor=center] at (med3) {1};
  \node[silver, anchor=center] at (med4) {2};
  \node[bronze, anchor=center] at (med2) {3};
  \node[anchor=center, inner sep=0pt] (robot_12) at (m12) {\includegraphics[height=5mm]{figures/bot1.png}};
  \node[anchor=center, inner sep=0pt] (robot_34) at (m34) {\includegraphics[height=5mm]{figures/bot1.png}};
  \node[anchor=center, inner sep=0pt] (robot_13) at (m13) {\includegraphics[height=5mm]{figures/bot1.png}};
  \node[anchor=center, inner sep=0pt] (robot_24) at (m24) {\includegraphics[height=5mm]{figures/bot1.png}};
  \node[anchor=center, inner sep=0pt] (robot_14) at (m14) {\includegraphics[height=5mm]{figures/bot1.png}};
  \node[anchor=center, inner sep=0pt] (robot_23) at (m23) {\includegraphics[height=5mm]{figures/bot1.png}};

  \node[bubble, callout absolute pointer={(robot_13.east)}]
  at ([shift={(1.0,0)}]robot_13.east) {$y_3$ wins!};

  \node[sworst]  (s2_1) at ([shift={(\acColA,\acScoreY)}]P2.north) {$r_1=0$};
  \node[sbox] (s2_2) at ([shift={(\acColB,\acScoreY)}]P2.north) {$r_2=1$};
  \node[sbest]   (s2_3) at ([shift={(\acColC,\acScoreY)}]P2.north) {$r_3=3$};
  \node[sbox]   (s2_4) at ([shift={(\acColD,\acScoreY)}]P2.north) {$r_4=2$};

  \node[calloutpos] (C2) at ([yshift=2mm]P2.south)
  {quality comparison\,$\rightarrow$\,{\fontsize{7.5}{9}\selectfont\textbf{win counts}}};

  \node[panel] (P3) at (11.6,0) {};
  \node[pnum] at ([shift={(4mm,-4mm)}]P3.north west) {3};
  \node[ptitle, anchor=north] (P3t) at ([yshift=-3mm]P3.north) {Self-Uncertainty};
  \node[psub,   anchor=north] (P3s) at ([yshift=-0.5mm]P3t.south)
  {inverse average per-token entropy: $1/\bar H(y_i)$};

  \draw[gray!35, rounded corners=3pt]
  ([shift={(-2.35,-1.25)}]P3.north) rectangle ([shift={(2.60,-3.55)}]P3.north);

  \node[ylab] (y1l) at ([shift={(\acYlabX,-1.6)}]P3.north) {$y_1$:};
  \node[htok, heat=0.19, anchor=base west] (t1a)
  at ([xshift=2pt]y1l.base east) {95};

  \node[ylab] (y2l) at ([shift={(\acYlabX,-2.15)}]P3.north) {$y_2$:};
  \node[htok, heat=0.16, anchor=base west] (t2a)
  at ([xshift=2pt]y2l.base east) {The};
  \node[htok, heat=0.45, anchor=base west] (t2b)
  at ([xshift=1pt]t2a.base east) {\ result};
  \node[htok, heat=0.90, anchor=base west] (t2bp)
  at ([xshift=1pt]t2b.base east) {\ probably};
  \node[htok, heat=0.55, anchor=base west] (t2c)
  at ([xshift=1pt]t2bp.base east) {\ is};
  \node[htok, heat=0.75, anchor=base west] (t2d)
  at ([xshift=1pt]t2c.base east) {\ 95};

  \node[ylab] (y3l) at ([shift={(\acYlabX,-2.75)}]P3.north) {$y_3$:};
  \node[htok, heat=0.45, anchor=base west] (t3a)
  at ([xshift=2pt]y3l.base east) {83};
  \node[htok, heat=0.08, anchor=base west] (t3b)
  at ([xshift=1pt]t3a.base east) {+};
  \node[htok, heat=0.15, anchor=base west] (t3c)
  at ([xshift=1pt]t3b.base east) {27};
  \node[htok, heat=0.05, anchor=base west] (t3d)
  at ([xshift=1pt]t3c.base east) {=};
  \node[htok, heat=0.20, anchor=base west] (t3e)
  at ([xshift=1pt]t3d.base east) {110};
  \node[htok, heat=0.15, anchor=base west] (t3f)
  at ([xshift=1pt]t3e.base east) {;};
  \node[htok, heat=0.15, anchor=base west] (t3g)
  at ([xshift=1pt]t3f.base east) {110};
  \node[htok, heat=0.05, anchor=base west] (t3h)
  at ([xshift=1pt]t3g.base east) {-};
  \node[htok, heat=0.15, anchor=base west] (t3i)
  at ([xshift=1pt]t3h.base east) {15};
  \node[htok, heat=0.05, anchor=base west] (t3j)
  at ([xshift=1pt]t3i.base east) {=};
  \node[htok, heat=0.18, anchor=base west] (t3k)
  at ([xshift=1pt]t3j.base east) {95};

  \node[ylab] (y4l) at ([shift={(\acYlabX,-3.3)}]P3.north) {$y_4$:};
  \node[htok, heat=0.16, anchor=base west] (t4a)
  at ([xshift=2pt]y4l.base east) {The};
  \node[htok, heat=0.28, anchor=base west] (t4b)
  at ([xshift=1pt]t4a.base east) {\ answer};
  \node[htok, heat=0.32, anchor=base west] (t4c)
  at ([xshift=1pt]t4b.base east) {\ is};
  \node[htok, heat=0.30, anchor=base west] (t4d)
  at ([xshift=1pt]t4c.base east) {\ 95};

  \tikzset{
    hbadge/.style={
      font=\tiny, anchor=base west, inner xsep=2pt, inner ysep=1pt,
      draw=plotblue!50, rounded corners=2pt, fill=plotblue!8, text=black,
    },
  }
  \foreach \col/\cx/\sVal/\bg/\border in {%
    1/\acColA/5.3/{plotblue!7}/{plotblue!45},
    2/\acColB/1.8/{plotred!10}/{plotred!55},
    3/\acColC/6.7/{plotgreen!22}/{plotgreen!75},
    4/\acColD/3.7/{plotblue!7}/{plotblue!45}%
  } {
    \node[tcell, fill=\bg, anchor=center, draw=\border, inner xsep=3pt, inner ysep=2pt,
      font=\scriptsize, rounded corners=3pt,
    minimum width=9mm, minimum height=4mm]
    (tabB\col) at ([shift={(\cx,\acScoreY)}]P3.north) {$r_\col=\sVal$};
  }

  \node[calloutpos] (C3) at ([yshift=2mm]P3.south)
  {token entropy\,$\rightarrow$\,{\fontsize{7.5}{9}\selectfont\textbf{confidence scores}}};
\end{tikzpicture}

%% file: sections/related_work.tex
\section{Related Work}\label{sec:related-work}

\paragraph{Easy-to-hard generalization and data efficiency}
Training on easy examples can transfer surprisingly well to harder tasks.
SFT on easy data can be competitive with models trained on hard data \citep{haseUnreasonableEffectivenessEasy2024}, and a reward model trained on easy data can supervise hard data training \citep{sun2024easytohardgeneralizationscalablealignment}. Relatedly, labels from a weak model can elicit stronger capabilities in a strong learner \citep{burns2023weaktostronggeneralizationelicitingstrong}.
Finally, \citet{wangReinforcementLearningReasoning2025} show that continued RL on a single example improves performance on MATH500, though most of the gain is attributed to format correction.
We take these works one step further by focusing on fully saturated training data, rather than just easy data where the model is strong but not perfect.

\paragraph{Reward signals}
The choice of reward signal is critical in post-training, and several recent works have explored alternatives to correctness-based rewards. When verifiable rewards are unavailable, dedicated outcome reward models (ORMs) are used to judge the complete trace, process reward models provide step-level supervision \citep{zhengSurveyProcessReward2025}, and rubric-based rewards \citep{huangReinforcementLearningRubric2025, gunjalRubricsRewardsReinforcement2025} provide structured criteria for subjective outputs, but can be tedious to construct. Both alternative metrics we study are outcome-level signals.

\paragraph{Self-improvement}
One of our metrics explicitly uses the model itself as a judge, leading to a process of self-improvement, which has been explored extensively in prior work, e.g., by using majority voting \citep{shafayatCanLargeReasoning2025}, multi-role self-play \citep{yangSPELLSelfPlayReinforcement2025}, meta-judge training \citep{wuMetaRewardingLanguageModels2025}, or token-probability self-judging \citep{gargIPOYourLanguage2025}. However, so far, none of these works has focused on the regime of saturated training data.

\paragraph{Best-of-n selection}
Sampling multiple completions and selecting the best one is a general technique for improving model performance~\citep{nakanoWebGPTBrowserassistedQuestionanswering2022}, which has been adapted for online SFT training~\citep{dongRAFTRewardRAnked2023a}.
Several recent methods extend this to full rankings \citep{niuRankingRewardIntraGroup2025, choiGOPOPolicyOptimization2026, songPreferenceRankingOptimization2024}.
However, these methods have not yet been applied to the regime of saturated training data.

\paragraph{Entropy as signal}
Predictive uncertainty over generated tokens has been proposed in prior work as a proxy for output quality, including token-level entropy for translation quality estimation \citep{fomichevaUnsupervisedQualityEstimation2020a}, semantic entropy for free-form QA correctness \citep{kuhnSemanticUncertaintyLinguistic2023}, and epistemic uncertainty for hallucination detection \citep{xiaoHallucinationPredictiveUncertainty2021}. Beyond its use as a passive signal, entropy minimization is an effective training objective by itself, improving reasoning performance without any labels \citep{agarwalUnreasonableEffectivenessEntropy2025}.

%% file: sections/methods.tex
\section{Methods}\label{sec:part2-methodology}

We develop a methodology for extracting a training signal from saturated questions and investigate whether the resulting improvements transfer to hard questions the model cannot yet reliably solve. We first partition questions by difficulty to isolate saturated-to-hard transfer (\cref{sec:identifying-trainingset}), then introduce two quality signals for ranking correct completions (\cref{sec:scoring-methods}), and finally apply them within DPO (\cref{sec:methods-dpo}) and a logistic-weighted RRHF variant (\cref{sec:rrhf}).

\paragraph{Preliminaries}
An LLM $M$ is characterized by a probability distribution $\pi_M$ over sequences of tokens in a vocabulary $\mathcal{V}$, generating completions $y$ from prompts $x$.
A reward function $R(x, y) \in \mathbb{R}$ scores completion quality. We sample $n$ completions per problem and write $\{y_i\}_{i=1}^n$ for this group, with $r_i = R(x, y_i)$ for their rewards. The model's uncertainty at each step is captured by the \emph{token-level entropy}:
\begin{equation}
  H_t(x, y_{<t}) = -\sum_{v \in \mathcal{V}} \pi_M(v \mid x, y_{<t}) \log \pi_M(v \mid x, y_{<t}).
  \label{eq:token-entropy-methods}
\end{equation}

\subsection{Difficulty Partitioning}\label{sec:identifying-trainingset}

We consider two related datasets: $\mathcal{D}$, the target distribution of interest, and $\mathcal{D}'$, which is drawn from a strictly simpler distribution than $\mathcal{D}$. For instance, $\mathcal{D}$ could represent research-level mathematics problems, while $\mathcal{D}'$ could represent high-school-level ones. We estimate the model's per-question solve rate by the empirical accuracy over $n$ samples, $\hat{p}(x) = \tfrac{1}{n}\sum_i r_i$, and define two saturated splits:
\begin{align*}
   \Qeasy &= \{x \in \mathcal{D} : \hat{p}(x) = 1\} \text{ and }\\
   \Qstrict &= \{x \in \mathcal{D}' : \hat{p}(x) = 1\}.
\end{align*}
This distinction allows us to isolate saturated-to-hard transfer and makes it easier to analyze the relationship between training signals and accuracy improvements. The question we seek to answer is whether we can improve accuracy on $\mathcal{D} \setminus \Qeasy$, solely with access to $\Qstrict$.

\subsection{Quality Scoring}\label{sec:scoring-methods}

Having filtered for saturated questions, training algorithms additionally require a signal to optimize. Since all completions are (by definition) correct, we need a quality signal to differentiate them and construct training pairs. We consider two complementary signals: pairwise LLM judgments from the model itself or a stronger teacher, and token-level entropy as a quality heuristic.

\subsubsection{LLM-as-a-judge}

We define a \emph{judge} as a function $J(x, y_i, y_j) \in \{0, 1\}$ that returns 1 if $y_i \succ y_j$ and 0 otherwise. Each unordered pair $\{y_i, y_j\}$ is compared exactly once. The \emph{win-rate score} of completion $y_i$ is its fraction of won comparisons:
\begin{equation}
  r_i^J = \frac{1}{n-1} \sum_{j \neq i} \mathbf{1}[y_i \succ y_j].
  \label{eq:win-rate}
\end{equation}
When $J = J_M$ is implemented by the policy model $\pi_M$ itself, we call this \emph{self-judging}.
The prompts used are in \cref{app:prompts}.

\subsubsection{Self-Uncertainty}
Prior work has established predictive uncertainty over generated tokens (including token-level entropy and related semantic and epistemic measures) as a reliable proxy for output quality \citep{fomichevaUnsupervisedQualityEstimation2020a,kuhnSemanticUncertaintyLinguistic2023,xiaoHallucinationPredictiveUncertainty2021}.
We exploit this as a judge-free alternative to pairwise LLM evaluation. We define the inverse entropy score as:
\begin{equation*}
  r_i^H = 1/\max(\bar{H}(x, y_i), \epsilon),
\end{equation*}
where $\bar{H}(x, y_i) = \frac{1}{T}\sum_{t=1}^T H_t(x, y_{i,<t})$ is the mean token-level entropy of completion $y_i$ (\cref{eq:token-entropy-methods}), a scalar measure of how deterministic the response is, and a small $\epsilon$ guards against division by zero. Post-training and instruction-tuning systematically lower token-level entropy~\citep{yuDAPOOpenSourceLLM2025,cao2026entropycalibrationlanguagemodels}, so selecting low-entropy completions from a base model may approximate the structured distribution instruction-tuning would induce.

\subsection{DPO on Saturated Questions}\label{sec:methods-dpo}


DPO requires a preference pair, a single chosen and a single rejected completion, per question for training. However, on $\Qeasy$, correctness can no longer distinguish the chosen completion from the rejected one. Instead, these pairs are constructed purely from quality scores, $y_w = \arg\max_{y_i} r_i$ and $y_l = \arg\min_{y_i} r_i$,\footnote{Ties at the top or bottom are broken by random selection; if all scores are equal, the question is skipped.} where $r_i$ is the score assigned by the selected quality signal.

\subsection{\texorpdfstring{Logistic-Weighted RRHF ($\sigma$-RRHF)}{Logistic-Weighted RRHF (sigma-RRHF)}}\label{sec:rrhf}

RRHF~\citep{yuanRRHFRankResponses2023} aligns model probabilities with reward rankings via a hinge-style loss over response pairs. In particular, given $n$ completions $\{y_i\}_{i=1}^n$, the method decomposes the loss into two parts. First, a \emph{rank loss} penalizes pairs where a lower-reward response receives higher model probability. Formally, the method assigns each response a length-normalized log-probability score $p_i = \frac{\log \pi_M(y_i \mid x)}{|y_i|}$ and defines the rank loss as:
\begin{equation}
  \mathcal{L}_{\text{rank}} = \sum_{r_i < r_j} \max(0,\, p_i - p_j).
  \label{eq:rank-loss-methods}
\end{equation}

Second, an \emph{SFT loss} anchors the model to the highest-reward response, $y_{i'}$ with $i' = \arg\max_i r_i$:
\begin{equation}
  \mathcal{L}_{\text{ft}} = -\log \pi_M(y_{i'} \mid x).
  \label{eq:sft-loss-methods}
\end{equation}
The total RRHF loss is $\mathcal{L} = \mathcal{L}_{\text{rank}} + \mathcal{L}_{\text{ft}}$.
To apply RRHF to our setting, we directly optimize the model to align its probabilities with the quality-based rankings induced by the selected quality signal.

\paragraph{Logistic weighting}
In the vanilla rank loss of \cref{eq:rank-loss-methods}, the penalty for a violated pair is determined solely by the log-probability margin $\max(0,\, p_i - p_j)$, regardless of how far apart the two responses are in quality. We add a multiplicative weight equal to the logistic sigmoid of the score gap:
\begin{equation*}
  \mathcal{L}_{\text{rank}}^{\sigma} = \sum_{r_i < r_j} \sigma(r_j - r_i)\,\max(0,\, p_i - p_j),
\end{equation*}
where $\sigma$ is the logistic sigmoid. When two responses have nearly identical quality scores, the penalty is halved. When the gap is large, the weight approaches~1.
The total loss combines the weighted rank loss with the SFT anchor on the best response:
\begin{equation*}
  \mathcal{L} = \lambda\,\mathcal{L}_{\text{rank}}^{\sigma} + \mathcal{L}_{\text{ft}},
\end{equation*}
where the rank loss weight $\lambda$ balances the ranking signal against the SFT regularization. In standard RRHF, both terms are weighted equally ($\lambda = 1$).


%% file: sections/results.tex
\section{Results}\label{sec:results}
We evaluate the methods developed previously on saturated \texttt{chain sum} questions (\cref{sec:results-easy}). We further test generalization to GSM8K and find that the findings transfer only partially (\cref{sec:gsm8k}). Finally, we apply the methods to an instruction-tuned model to test limitations of entropy-based scoring.

\subsection{Experimental Setup}
\paragraph{Task and data} We use the \texttt{chain sum} task from ReasoningGym~\citep{stojanovskiREASONINGGYMReasoning2025}, in which a sequence of additions and subtractions must be evaluated left to right. A representative instance is shown in \cref{app:task-example}. Task difficulty is controlled via two parameters: the number of terms and the number of digits per operand. 

We use \usedmodel{}~\citep{yangQwen3TechnicalReport2025} as our base model throughout all experiments. All methods share the same generation pool: \texttt{chain sum} problems with terms and digits ranging from 6 to 10, each with $n=8$ sampled generations from the base model. From this we construct the saturated split $\Qeasy$ as defined in \cref{sec:identifying-trainingset}. For $\Qstrict$, we use problems with terms and digits ranging from 3 to 6, excluding 6/6.
Further details on evaluation protocol and training are in \cref{app:hyperparameters}.

To probe whether quality differences matter beyond chance, we additionally include a random baseline that assigns each completion a score uniformly at random. We refer to this baseline, self-judge, strong judge, and inverse entropy as \emph{scoring methods}, and instantiate the algorithms with each where applicable.

\subsection{Learning from \emph{Saturated} Questions}\label{sec:results-easy}
We train on questions the model already solves consistently, first on $\Qeasy$ and then on the stricter $\Qstrict$ partition. For both splits, we compare self-judge and inverse-entropy scoring, and include \strongmodel{} as a stronger external judge on $\Qstrict$ to contextualize the quality of these signals. The results are presented in \cref{tab:results-easy-combined}.

\input{figures/results_easy_strictly_easy.tex}

\paragraph{Methods on saturated questions}
On $\Qeasy$, all training methods improve over the base model in at least one metric, but the gains depend strongly on the scoring signal. SFT already improves substantially over the base model, making it the natural reference point for methods that claim to extract additional signal from saturated data. This separates two effects: saturated examples are useful even under ordinary imitation, but quality-ranked methods must show gains beyond copying accepted correct completions.
Self-judged DPO is comparatively weak: it improves over the base model, but remains below SFT. Inverse-entropy DPO mainly helps $\text{pass}@8$, while the clearest benefit from quality ranking appears in $\sigma$-RRHF. With self-judge scoring, $\sigma$-RRHF exceeds SFT on $\text{pass}@1$ and roughly matches its $\text{pass}@8$; with inverse entropy, it achieves the strongest result: $29.25\%$ $\text{pass}@1$ and $61.75\%$ $\text{pass}@8$.
The strength of inverse entropy is surprising because it requires no external evaluator. One possible explanation is model compatibility: low-entropy completions may be solutions the model can already represent stably, so training on them reinforces robust computations rather than pushing the model toward preferences defined outside its own distribution.

\paragraph{Strictly saturated questions}
We extend the analysis to $\Qstrict$, where questions are intrinsically simpler by construction. This tests a stronger form of transfer: the training questions are not only saturated, but also easier than the evaluation distribution. SFT still improves $\text{pass}@1$ over the base model, confirming that very easy saturated examples contain reusable algorithmic signal. The smaller gains relative to $\Qeasy$ suggest that part of the benefit comes from distributional overlap with the harder evaluation problems.

Under self-judge scoring, DPO changes very little between $\Qeasy$ and $\Qstrict$ and remains below SFT on $\text{pass}@1$; we refer to \cref{app:dpo-hyperparam} for a possible explanation. In contrast, $\sigma$-RRHF is more sensitive to the quality signal: it improves over SFT with self-judge scoring on both metrics, and inverse entropy gives the strongest $\Qstrict$ result.

To compare against a stronger external quality signal on $\Qstrict$, we replace the self-judge with \strongmodel{}, a stronger instruction-tuned model. This improves over the self-judged $\sigma$-RRHF row ($19.69\%$ $\text{pass}@1$, $58.00\%$ $\text{pass}@8$). However, the strongest $\Qstrict$ result surprisingly comes from $\sigma$-RRHF with inverse entropy ($22.35\%$ $\text{pass}@1$, $59.25\%$ $\text{pass}@8$).
This is consistent with the model-compatibility explanation above: inverse entropy can outperform the strong judge despite using only the base model's token probabilities.

\subsection{GSM8K Experiments}\label{sec:gsm8k}
To test whether our findings transfer to another task, we repeat key experiments on GSM8K~\citep{cobbeTrainingVerifiersSolve2021}. We construct $\Qeasy$ using the same mechanism as for \texttt{chain sum} (\cref{sec:identifying-trainingset}), yielding $|\Qeasy| = 2{,}302$ questions ($\hat{p}(x)=1$, estimated for \usedmodel). We evaluate on the full GSM8K test set.

\paragraph{Methods on GSM8K}
We compare DPO and $\sigma$-RRHF on $\Qeasy$ from GSM8K.
We compare the same scoring family used above: random scoring, the self-judge, a strong judge, and inverse entropy (\cref{tab:gsm8k-combined}).

For DPO, downstream accuracy improves by $+7.7\%$ for the strong judge, $+3.3\%$ for inverse entropy, $+1.5\%$ for self-judge, and $-0.1\%$ for random scoring. $\sigma$-RRHF with inverse entropy and the strong judge both improve $\text{pass}@1$ while preserving $\text{pass}@8$, but the self-judge degrades performance well below the base model. Overall, $\sigma$-RRHF underperforms DPO across all scoring methods, plausibly because DPO's implicit KL constraint limits divergence from the reference model, whereas $\sigma$-RRHF's only regularization is its SFT anchor.
\input{figures/gsm8k_combined_eval_v2.tex}

\paragraph{Scorer agreement on GSM8K}\label{sec:gsm8k-scoring}
To investigate why the performance improvements are more modest on GSM8K, we measure how closely each scoring method agrees with the strong judge via within-question Spearman rank correlation, aggregated over $\Qeasy$.
On GSM8K, we find moderate positive correlation between inverse entropy and the strong judge ($\rho = 0.56$), supporting entropy as a selection criterion. In contrast, the self-judge and strong judge slightly disagree on GSM8K ($\rho = {-}0.2$), compared to a weak positive correlation on \texttt{chain sum} ($\rho = 0.19$).

This explains why the model trained with the self-judge metric underperforms even the base model: the self-judge provides actively misleading signal on GSM8K.
$\sigma$-RRHF is plausibly more affected than DPO because it fits the full ranking over all $n$ completions rather than only the best-worst pair, amplifying the systematic bias indicated by the negative self-judge/strong-judge Spearman correlation.

\subsection{Instruction-Tuned Models}\label{sec:results-instruct}
We test whether entropy-based DPO transfers to instruction-tuned models by repeating the GSM8K DPO experiments on the instruction-tuned variant of the same model family. As shown in \cref{tab:gsm8k-dpo-instruct-eval}, no method yields substantial improvements over the base model, and inverse entropy DPO even degrades performance. This contrasts with the base model, where inverse entropy DPO improved performance significantly. This is likely due to the low-entropy bias of instruction-tuning: further training to lower entropy is much less likely to yield quality improvements and can instead over-optimize the proxy.

\input{figures/gsm8k_dpo_instruct_eval.tex}


%% file: figures/results_easy_strictly_easy.tex

\pgfplotstableread[col sep=comma]{data/chainsum_saturated_results.csv}\reseasycombined
\begin{table*}[t]
  \caption{
    Method comparison on saturated chain-sum problems with two training splits.
    \textbf{Bold} is best per column.
  }
  \label{tab:results-easy-combined}
  \centering
  \pgfplotstabletypeset[
    outfile=figures/pgf-static/results_easy_strictly_easy.tex,
    begin table=\begin{tabularx}{\textwidth},
    end table=\end{tabularx},
    col sep=comma,
    string type,
    columns={method, data, easy_pass1, easy_pass8, strict_pass1, strict_pass8},
    columns/method/.style={column name=Model / Training, column type=l},
    columns/data/.style={        column name=Scoring,             column type=X},
    columns/easy_pass1/.style={  column name=pass@1,           column type=r},
    columns/easy_pass8/.style={  column name=pass@8,           column type=r},
    columns/strict_pass1/.style={column name=pass@1,           column type=r},
    columns/strict_pass8/.style={column name=pass@8,           column type=r},
    every head row/.style={
      before row={%
        \toprule
        & & \multicolumn{2}{c}{$\Qeasy$} & \multicolumn{2}{c}{$\Qstrict$} \\
        \cmidrule(lr){3-4}\cmidrule(lr){5-6}
      },
      after row=\midrule,
    },
    every last row/.style={after row=\bottomrule},
    every row no 0/.style={after row={\arrayrulecolor{black!20}\midrule\arrayrulecolor{black}}},
    every row 5 column 2/.style={postproc cell content/.append style={/pgfplots/table/@cell content/.add={\bfseries}{}}},
    every row 5 column 3/.style={postproc cell content/.append style={/pgfplots/table/@cell content/.add={\bfseries}{}}},
    every row 5 column 4/.style={postproc cell content/.append style={/pgfplots/table/@cell content/.add={\bfseries}{}}},
    every row 5 column 5/.style={postproc cell content/.append style={/pgfplots/table/@cell content/.add={\bfseries}{}}},
  ]{\reseasycombined}
\end{table*}

%% file: figures/gsm8k_combined_eval_v2.tex

\pgfplotstableread[col sep=comma]{data/gsm8k_main_results.csv}\gsmcombinedv
\begin{table}[t]
  \caption{
    GSM8K accuracy for DPO and $\sigma$-RRHF variants trained on $\Qeasy$,
    evaluated on the test set.
    Inv.\ Entropy = inverse entropy.
    All $\sigma$-RRHF runs use $\lambda = 0.1$.
    \textbf{Bold} is best per column.
  }
  \label{tab:gsm8k-combined}
  \centering
  \setlength{\tabcolsep}{4pt}
  \pgfplotstabletypeset[
    outfile=figures/pgf-static/gsm8k_combined_eval_v2.tex,
    begin table=
    \begin{tabularx}{\columnwidth},
      end table=
    \end{tabularx},
    col sep=comma,
    string type,
    columns={method, scoring, pass1, pass8},
    columns/scoring/.style={column name=Scoring, column type=X},
    columns/method/.style={column name=Model / Training, column type=l},
    columns/pass1/.style={column name=pass@1, column type=r},
    columns/pass8/.style={column name=pass@8, column type=r},
    every head row/.style={before row=\toprule, after row=\midrule},
    every last row/.style={after row=\bottomrule},
    every row no 1/.style={after row=\midrule},
    every row no 3/.style={after row={\arrayrulecolor{black!20}\midrule\arrayrulecolor{black}}},
    every row no 5/.style={after row={\arrayrulecolor{black!20}\midrule\arrayrulecolor{black}}},
    every row no 7/.style={after row={\arrayrulecolor{black!20}\midrule\arrayrulecolor{black}}},
    every row 2 column 2/.style={postproc cell content/.append style={/pgfplots/table/@cell content/.add={\bfseries}{}}},
    every row 2 column 3/.style={postproc cell content/.append style={/pgfplots/table/@cell content/.add={\bfseries}{}}},
  ]{\gsmcombinedv}
\end{table}

%% file: figures/gsm8k_dpo_instruct_eval.tex

\pgfplotstableread[col sep=comma]{data/gsm8k_instruct_dpo_results.csv}\gsmdpoinstructtable
\begin{table}[tb]
  \caption{
    GSM8K accuracy for DPO variants starting from \texttt{Qwen3-1.7B}.
    \mbox{+ DPO} pairs are constructed from quality scores on saturated questions ($\Qeasy$);
    \mbox{+ DPO (correctness)} instead uses correct-vs-incorrect pairs from hard questions.
    Inv.\ Entropy = inverse entropy.
  }
  \label{tab:gsm8k-dpo-instruct-eval}
  \centering
  \setlength{\tabcolsep}{4pt}
  \pgfplotstabletypeset[
    outfile=figures/pgf-static/gsm8k_dpo_instruct_eval.tex,
    begin table=\begin{tabularx}{\columnwidth},
    end table=\end{tabularx},
    col sep=comma,
    string type,
    columns={method, data, pass1, pass8},
    columns/method/.style={column name=Model / Training, column type=l},
    columns/data/.style={column name=Scoring,               column type=X},
    columns/pass1/.style={column name=pass@1,             column type=r},
    columns/pass8/.style={column name=pass@8,             column type=r},
    every head row/.style={before row=\toprule, after row=\midrule},
    every last row/.style={after row=\bottomrule},
    every row no 0/.style={after row={\arrayrulecolor{black!20}\midrule\arrayrulecolor{black}}},
  ]{\gsmdpoinstructtable}
\end{table}

%% file: sections/discussion.tex
\section{Limitations and Future Work}
\label{sec:discussion}

We briefly discuss three limitations of our work and directions for future research.

First, our experiments use a single model family (\usedmodel{}) and reveal that the relative effectiveness of training methods does not transfer straightforwardly across domains: $\sigma$-RRHF dominates on \texttt{chain sum}'s constrained output space, while DPO outperforms it on GSM8K's noisier free-form reasoning. Investigating these dynamics on larger model scales and tasks is left as future work.


Second, we focus on only two quality metrics, self-judging and inverse entropy, but there are many other potential metrics that could be explored. For instance, rubric-based rewards~\citep{huangReinforcementLearningRubric2025, gunjalRubricsRewardsReinforcement2025} could be adapted to this setting by using the rubric as a quality signal rather than a reward model. More generally, investigating the relationship between metric reliability and training efficacy is an important direction for future work.

Finally, we have applied quality-based scoring only to offline methods, but it could also be integrated into RL algorithms. For instance, quality-based scoring could be directly applied to GRPO \citep{shaoDeepSeekMathPushingLimits2024}. However, RL is notoriously sensitive to misspecification and expensive to run, so careful experimentation would be needed to determine whether the stronger signal from quality-based scoring outweighs the instability in this setting, which we leave as future work.

%% file: sections/conclusion.tex
\section{Conclusion}
\label{sec:conclusion}

We study post-training on saturated questions, where correctness rewards no longer distinguish among completions. Our results show that quality differences among correct solutions can still provide a useful training signal and transfer to harder questions, but only when the quality scorer is reliable. This makes scorer quality the main bottleneck for learning from saturated data.

%% file: sections/impact_statement.tex
\section*{Impact Statement}

This work studies how to improve language models using quality signals from questions they already answer correctly. Such methods could make existing datasets more useful and reduce the need for additional annotation. The main risks are that unreliable self-judgments may reinforce model biases, while entropy-based objectives may reduce solution diversity or increase overconfidence. These risks make scorer calibration and held-out validation important when applying quality-based training.

%% file: sections/acknowledgements.tex
\section*{Acknowledgements}

HH used compute from the Swiss AI Initiative supported by a grant from the Swiss National Supercomputing Centre (CSCS) under project ID a155 on Alps.

%% file: sections/appendix.tex
\clearpage
\appendix
\onecolumn
\section{Task and Data Details}\label{app:details}
This appendix collects ablation studies, hyperparameter sweeps, and auxiliary analyses that support the main results.
\subsection{Difficulty Partitioning: Formal Setup}\label{app:difficulty-partitioning}

\paragraph{Problem setting}
Let questions $x$ be drawn from some distribution over verifiable problems with binary reward. We assume access to a finite collection $\mathcal{D}$ of such questions, as well as a second collection $\mathcal{D}'$ that is strictly simpler than $\mathcal{D}$ under some complexity measure $c$ (e.g.\ number of reasoning steps or operand magnitude). As intuition, consider $\mathcal{D}$ as research-level competition problems and $\mathcal{D}'$ as high-school-level ones; our concrete instantiation operates at a smaller scale, but the underlying principle is the same. We estimate the model's per-question solve rate by the empirical $\text{pass}@1$ over $n$ samples:
\begin{equation*}
  \hat{p}(x) = \frac{1}{n}\sum_{i=1}^n r_i.
\end{equation*}

\paragraph{Motivation}
Standard preference-based post-training requires both correct and incorrect responses per question to form preference pairs. Hard questions offer the most room for improvement but yield few correct completions, and those that are correct may not reflect sound reasoning. Saturated questions offer the opposite trade-off: dense, high-quality correct completions but no obvious room for direct improvement. We therefore partition the training set by empirical solve rate, separating easy from hard questions so that any observed transfer from easy-question training to hard-question performance can be cleanly attributed.

\paragraph{Partitioning criteria}
We partition $\mathcal{D}$ into two subsets based on the empirical solve rate $\hat{p}(x)$, and define a third subset from $\mathcal{D}'$:
\begin{align*}
  \Qeasy   &= \bigl\{x \in \mathcal{D}  : \hat{p}(x) = 1\bigr\} \\
  \Qhard   &= \bigl\{x \in \mathcal{D}  : \hat{p}(x) \leq \tfrac{1}{4}\bigr\} \\
  \Qstrict &= \bigl\{x \in \mathcal{D}' : \hat{p}(x) = 1\bigr\}
\end{align*}
Questions between the two thresholds ($\tfrac{1}{4} < \hat{p}(x) < 1$) are excluded from training splits so that observed performance changes are less likely to be driven by mixed difficulty levels in the training data. Comparing transfer from $\Qstrict$ versus $\Qeasy$ to hard-question performance probes the effect of the difficulty gap between training and evaluation data.

\subsection{Breaking Down Task Difficulty}\label{subsec:chainsum_breakdown}\label{app:pass-at-1}
The empirical solve rates used to define the difficulty partitions are shown in \cref{tab:pass-at-1}: performance decreases monotonically as the number of terms and digits increases, giving a controlled difficulty axis for the \texttt{chain sum} task.
\input{figures/pass_at_1_table}
\FloatBarrier

We further investigate whether arithmetic properties beyond the number of terms and digits explain difficulty variation within a single difficulty cell.

Within a fixed difficulty cell (same number of terms and digits), one might expect finer-grained arithmetic properties to further explain variance in $\text{pass}@1$. We examine three such features: \\(1)~\emph{proportion of additions}: the number of addition operators out of all operators in the expression, since subtractions can drive intermediate values negative and may be harder to track; \\(2)~\emph{mean operand magnitude}: the average absolute value of the operands, as larger numbers require more carry operations; and \\(3)~\emph{maximum intermediate value}: the largest absolute value reached during left-to-right evaluation, which determines how large the numbers a model must reason about can grow mid-computation.\\ \cref{tab:chainsum_6_6_breakdown} shows that none of these features meaningfully stratifies difficulty within the 6-term/6-digit subgroup.

\begin{table}[H]
  \caption{$\text{pass}@1$ breakdown within the 6-term/6-digit subgroup ($n=2{,}955$) by three structural features. The $\text{pass}@1$ range across all bins is only $0.55$--$0.63$, indicating that difficulty is not explained by any of these arithmetic properties.}
  \label{tab:chainsum_6_6_breakdown}
  \centering
  \begin{tabular}{lrr lrrr lrrr}
    \toprule
    \multicolumn{3}{c}{\# additions (out of 5 ops)} &
    \multicolumn{4}{c}{Mean operand magnitude} &
    \multicolumn{4}{c}{Max intermediate value} \\
    \cmidrule(r){1-3}\cmidrule(lr){4-7}\cmidrule(l){8-11}
    Bin & $n$ & $\text{pass}@1$ &
    Bin & $n$ & $\text{pass}@1$ & Mean &
    Bin & $n$ & $\text{pass}@1$ & Mean \\
    \midrule
    0 & 104 & 0.617 & Q1 & 591 & 0.585 & 401{,}274   & Q1 & 591 & 0.561 & 786{,}505     \\
    1 & 473 & 0.557 & Q2 & 591 & 0.557 & 493{,}138   & Q2 & 591 & 0.575 & 1{,}135{,}996 \\
    2 & 904 & 0.549 & Q3 & 591 & 0.555 & 550{,}571   & Q3 & 591 & 0.566 & 1{,}480{,}625 \\
    3 & 892 & 0.556 & Q4 & 591 & 0.550 & 608{,}392   & Q4 & 591 & 0.559 & 1{,}898{,}710 \\
    4 & 481 & 0.582 & Q5 & 591 & 0.568 & 700{,}397   & Q5 & 591 & 0.554 & 2{,}790{,}242 \\
    5 & 101 & 0.627 &    &     &       &             &    &     &       &               \\
    \bottomrule
  \end{tabular}%
\end{table}
\FloatBarrier

\subsection{Task Example}\label{app:task-example}
\input{figures/task_example}
{\small\textit{Both responses arrive at the correct answer but differ in verbosity and reasoning style.}}

\clearpage

\input{sections/hyperparameters.tex}

\clearpage

\section{Additional Experiments}\label{app:additional-results}

\subsection{Validating Methods on Hard Questions}\label{app:results-hard}
We first trained on $\Qhard$ (\cref{app:difficulty-partitioning}), where the model fails to solve most instances and binary correctness signal is abundant. We compare Rejection Sampling, DPO, and $\sigma$-RRHF. Rejection Sampling with one iteration serves as the primary baseline, which simplifies to SFT on correct completions only; $\sigma$-RRHF uses $\lambda{=}0.1$.

\paragraph{DPO with contrastive pairs} When training DPO on $\Qhard$, we filter out unsolvable training questions ($\text{pass}@8=0$) so that each question has at least one correct and one incorrect completion. Evaluation remains on the same held-out chain-sum evaluation set used in the saturated experiments. We consider two pair-construction strategies. \textbf{Random}: $y_w$ is sampled uniformly from the correct completions and $y_l$ from the incorrect ones, ignoring quality scores. \textbf{Self-judge}: $y_w = \arg\max_{i:\, r_i=1} r_i^{J_M}$ and $y_l = \arg\min_{i:\, r_i=0} r_i^{J_M}$, using the policy model's own win-rate scores (\cref{eq:win-rate}). Unless otherwise noted, DPO on $\Qhard$ uses the self-judge variant.

\cref{tab:results-hard} reveals a tension between per-sample accuracy and coverage. DPO achieves high $\text{pass}@1$ ($+4.8\%$ over SFT) but trails substantially on $\text{pass}@8$ ($-11\%$). SFT, which trains on all correct completions equally, preserves coverage but forgoes any quality distinction among them. $\sigma$-RRHF mitigates this trade-off: its SFT anchor preserves $\text{pass}@8$ ($+4.5\%$ over SFT) while its quality-weighted ranking provides the largest $\text{pass}@1$ gain ($+9.6\%$ over SFT).
\input{figures/results_hard.tex}

\paragraph{Random vs.\ self-judge pair selection}\label{app:results-randomvsjudge}
To isolate the effect of completion ranking, we compare the two pair-construction strategies above with all other variables fixed. To understand training dynamics, we additionally track the \emph{reward margin} (\cref{fig:dpo-reward-margins}), defined as the log-probability difference between the chosen and rejected completion under the current policy.

Self-judge pairs outperform random pairs on both metrics: $\text{pass}@1$ improves by $+5.4\%$ and $\text{pass}@8$ by $+1.5\%$ (\cref{tab:results-hard}). Reward margins grow throughout training for both strategies (\cref{fig:dpo-reward-margins}a), yet their downstream trajectories diverge: self-judge pairs continue improving $\text{pass}@1$ until $75\%$ of training, whereas random pairs plateau after $50\%$ (\cref{fig:dpo-reward-margins}b). The continued reward-margin growth for random pairs without corresponding evaluation gains suggests overfitting to the training signal. Self-judge pairs, by contrast, translate margin growth into downstream improvement for longer, suggesting that preference-signal quality strongly affects training efficacy. The pattern suggests that, on this split, the judge identifies more useful responses than random selection among completions with the same correctness level.
\input{figures/dpo_reward_margins.tex}

\subsection{Rank Loss Weight Ablation}\label{app:rank_weight}
\input{figures/rank_weight_eval.tex}
We ablate the rank loss weight $\lambda$ in RRHF on $\Qhard$, the hardest difficulty split, where the model has the most room for improvement and the ranking signal should matter most.
We sweep $\lambda \in \{0, 10^{-2}, 10^{-1}, 0.5, 1.0\}$. The baseline $\lambda=0$ corresponds to SFT on the judge-selected argmax completion only (no rank loss), which recovers RAFT~\citep{dongRAFTRewardRAnked2023}. We report the mean of three runs for $\lambda=0$.

The rank loss provides at most marginal improvement: $\text{pass}@1$ peaks at $\lambda=10^{-2}$ ($+1.1\%$ over no rank loss), while $\text{pass}@8$ peaks at $\lambda=10^{-1}$ ($+1.5\%$), suggesting a mild diversity benefit at moderate weights. In both cases $\lambda \geq 0.5$ actively hurts performance. This suggests that, in this setting, RRHF's performance is driven primarily by the SFT-on-argmax component (selecting and imitating the best completion via the judge) rather than the ranking objective itself. Because all completions in this ablation are ranked by the self-judge, whose agreement with the strong judge is weak (\cref{app:entropy}), it remains an open question whether a higher-quality ranker would unlock a larger benefit from the rank loss term.

\subsection{\texorpdfstring{$\sigma$-RRHF Component Ablation}{sigma-RRHF Component Ablation}}
To understand which components of $\sigma$-RRHF contribute to its performance, we ablate three design choices on $\Qhard$: the SFT term, the hinge in the rank loss, and the logistic pair reweighting.
\input{figures/rrhf_variants.tex}

\paragraph{No SFT loss} This ablation removes the SFT component entirely, training only with the rank loss. The total loss reduces to:
\begin{equation*}
  \mathcal{L} = \mathcal{L}_{\text{rank}}.
\end{equation*}
Without the SFT anchor on the best completion, the model receives only relative preference signal and has no direct imitation target.

\paragraph{No hinge} Instead of only penalizing incorrectly ranked pairs, we also reward correctly ranked ones by removing the $\max(0, \cdot)$ clamp. For every pair where $r_i < r_j$ (i.e.\ $y_j$ is the better completion), the loss becomes:
\begin{equation*}
  \mathcal{L}_{\text{rank}} = \sum_{r_i < r_j} \frac{1}{1+\exp(-(r_j - r_i))}\,(p_i - p_j).
\end{equation*}
This encourages the model to keep increasing the log-probability gap even for already correctly ranked pairs, which may lead to overconfident predictions.

\paragraph{No logistic weighting} This ablation removes the logistic reweighting, treating all pairs equally regardless of the reward gap. For every pair where $r_i < r_j$, the loss is:
\begin{equation*}
  \mathcal{L}_{\text{rank}} = \sum_{r_i < r_j} \max(0,\, p_i - p_j).
\end{equation*}

\paragraph{Results} We discuss the ablation results in \cref{tab:rrhf-variants}. Removing the SFT term is catastrophic: $\text{pass}@1$ drops from $39.75\%$ to $6.69\%$ and $\text{pass}@8$ from $77.50\%$ to $28.00\%$, showing that the rank loss alone is insufficient to guide learning in this setting. Removing the hinge costs roughly $3\%$ on both metrics ($36.56\%$ $\text{pass}@1$, $74.00\%$ $\text{pass}@8$), suggesting that continuing to push already correct rankings apart slightly hurts. Removing the logistic reweighting has negligible effect ($39.62\%$ $\text{pass}@1$, $77.00\%$ $\text{pass}@8$), indicating that weighting pairs by reward gap adds little when the self-judge scores are noisy. Together with the rank weight ablation (\cref{app:rank_weight}), these initial $\Qhard$ experiments suggest that RRHF's performance is driven mostly by the SFT-on-argmax component. We did not repeat the component ablation on $\Qeasy$ or $\Qstrict$, so this conclusion should not be interpreted as establishing the same decomposition for the saturated setting studied in the main experiments.

\subsection{DPO Hyperparameter Sensitivity}\label{app:dpo-hyperparam}
On both $\Qeasy$ and $\Qstrict$, self-judge DPO underperforms SFT on $\text{pass}@1$. A plausible explanation is that DPO's contrastive objective is more sensitive to scoring noise than SFT: when the self-judge cannot meaningfully separate chosen and rejected completions, training on potentially near-arbitrary preferences becomes counterproductive.
We sweep the KL penalty $\beta$ in DPO to find stable training configurations.
\input{figures/dpo_strict_sweep_eval.tex}
We sweep $\beta \in \{0.1, 0.3, 0.5, 2, 5\}$ on the $\Qstrict$ split with self-judge pair selection. The standard value $\beta=0.1$ suggested by \citet{rafailovDirectPreferenceOptimization2024} yields only $5.06\%$ $\text{pass}@1$, while stronger regularization substantially improves performance, with the best configuration ($\beta=5$, $\eta=5\times10^{-6}$, 4 epochs) reaching $13.56\%$ $\text{pass}@1$ --- still below SFT. This need for much stronger KL regularization is consistent with observations by \citet{gengDeltaLearningHypothesis2025,lambertTulu3Pushing2025} and likely reflects noisy self-judge preference pairs.

\subsection{GSM8K DPO Training Dynamics}\label{app:gsm8k-dpo-training}
We show the implicit reward margin $\hat{r}(x, y_w) - \hat{r}(x, y_l)$ over training on GSM8K for the DPO variants from \cref{sec:gsm8k}. Inverse entropy reaches the highest margin, the \strongmodel{} strong judge grows more steadily, and the Qwen3-1.7B-Base self-judge stays near zero --- consistent with its near-zero Spearman correlation with the strong judge.
\input{figures/gsm8k_dpo_training.tex}

\subsection{Scorer Agreement and Entropy Correlations}\label{app:entropy}

\begin{figure*}[tb]
  \centering
  \includegraphics[width=\textwidth]{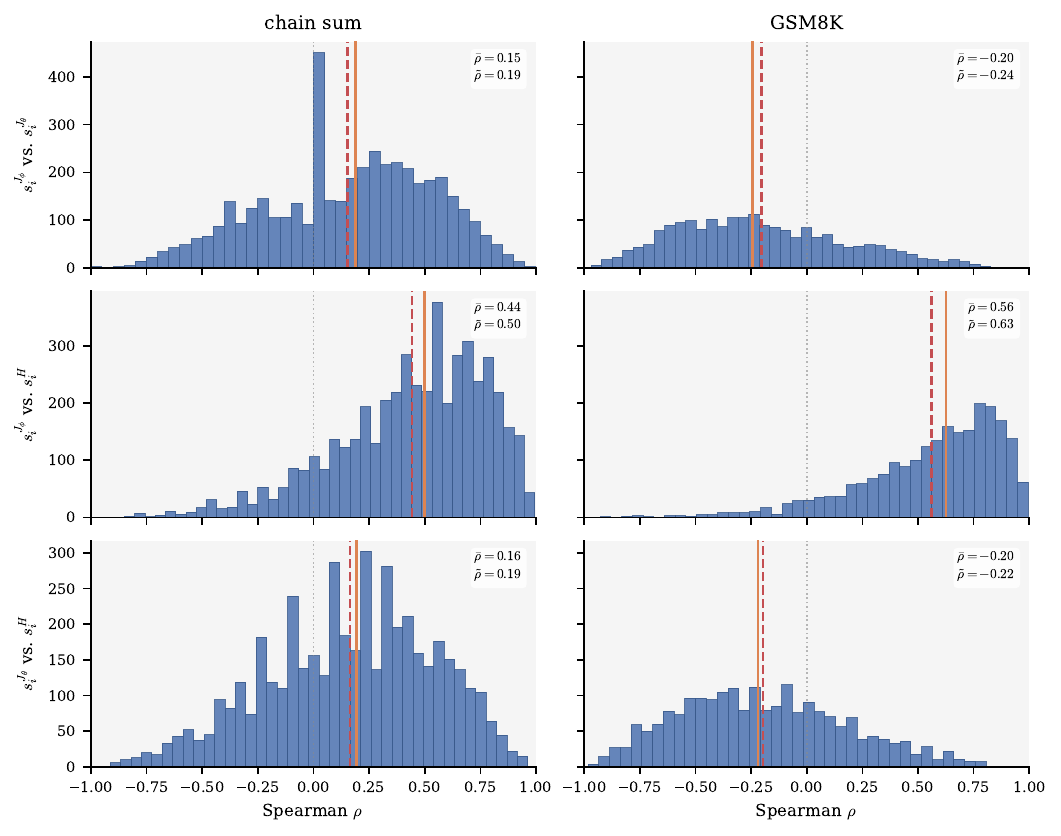}
  \caption{
    Distribution of within-question Spearman $\rho$ between each pair of scorers, for $\Qstrict$ \texttt{chain sum} (left) and $\Qeasy$ GSM8K (right). Row labels indicate the scorer pair. Dashed red line: mean ($\bar{\rho}$); solid orange line: median ($\tilde{\rho}$). The strong judge and inverse entropy show consistent positive correlation (middle row), while the self-judge agrees only weakly with the other two scorers.
  }
  \label{fig:judge-agreement}
\end{figure*}

We assess how well the three scoring methods agree on completion rankings: strong judge $J_\phi$ (\strongmodel), self-judge $J_\theta$ (Qwen3-1.7B-Base), and mean token-level entropy $\bar{H}$. For each question we rank its $n$ completions independently under each scorer and compute the within-question Spearman $\rho$ between every pair. \cref{tab:judge-agreement} reports the mean and median $\rho$ across questions; \cref{fig:judge-agreement} shows the full distributions. The \texttt{chain sum} correlations are computed on $\Qstrict$, where all questions are strictly easier by construction (terms and digits ranging from 3 to 6, excluding the 6-term/6-digit cell); the GSM8K correlations use $\Qeasy$.

The strong judge and inverse entropy correlate moderately on both tasks ($\bar{\rho} = +0.44$ on $\Qstrict$ \texttt{chain sum}, $+0.56$ on GSM8K), showing that lower-entropy completions tend to receive higher judge scores. The self-judge agrees only weakly with both the strong judge and inverse entropy on $\Qstrict$ \texttt{chain sum} ($\bar{\rho} \approx +0.15$ and $+0.16$), and on GSM8K the self-judge vs.\ inverse-entropy correlation reverses sign ($\bar{\rho} = -0.20$), consistent with the self-judge failures observed in \cref{sec:gsm8k}.

\input{figures/table_judge_agreement.tex}

\clearpage

\input{sections/prompts.tex}

%% file: figures/pass_at_1_table.tex
\newcommand{\passmin}{7}
\newcommand{\passmax}{93}

\begin{table}[H]
  \caption{Pass@1 (estimated from $k=8$ generations) for \usedmodel{} across chain-sum difficulty configurations. Cell color encodes pass@1 (darker = higher).}\label{tab:pass-at-1}
  \centering
  \begin{minipage}[t]{0.48\linewidth}
    \centering
    \begin{tabular}[t]{c|cccc}
      \toprule
      \shortstack{\emph{terms} \\ / \emph{digits}} & 3 & 4 & 5 & 6 \\
      \midrule
      3 & \heatcell{93}{\passmin}{\passmax} & \heatcell{91}{\passmin}{\passmax} & \heatcell{87}{\passmin}{\passmax} & \heatcell{82}{\passmin}{\passmax} \\
      4 & \heatcell{92}{\passmin}{\passmax} & \heatcell{87}{\passmin}{\passmax} & \heatcell{83}{\passmin}{\passmax} & \heatcell{79}{\passmin}{\passmax} \\
      5 & \heatcell{86}{\passmin}{\passmax} & \heatcell{81}{\passmin}{\passmax} & \heatcell{75}{\passmin}{\passmax} & \heatcell{68}{\passmin}{\passmax} \\
      6 & \heatcell{76}{\passmin}{\passmax} & \heatcell{70}{\passmin}{\passmax} & \heatcell{63}{\passmin}{\passmax} & \heatcell{56}{\passmin}{\passmax} \\
      \bottomrule
    \end{tabular}
    \par\smallskip
    {\small (a) $\Qstrict$ (3--6 terms \& digits)}
  \end{minipage}\hfill
  \begin{minipage}[t]{0.48\linewidth}
    \centering
    \begin{tabular}[t]{c|ccccc}
      \toprule
      \shortstack{\emph{terms} \\ / \emph{digits}} & 6 & 7 & 8 & 9 & 10 \\
      \midrule
      6  & \heatcell{56}{\passmin}{\passmax} & \heatcell{50}{\passmin}{\passmax} & \heatcell{43}{\passmin}{\passmax} & \heatcell{32}{\passmin}{\passmax} & \heatcell{27}{\passmin}{\passmax} \\
      7  & \heatcell{47}{\passmin}{\passmax} & \heatcell{41}{\passmin}{\passmax} & \heatcell{34}{\passmin}{\passmax} & \heatcell{25}{\passmin}{\passmax} & \heatcell{18}{\passmin}{\passmax} \\
      8  & \heatcell{40}{\passmin}{\passmax} & \heatcell{33}{\passmin}{\passmax} & \heatcell{27}{\passmin}{\passmax} & \heatcell{19}{\passmin}{\passmax} & \heatcell{14}{\passmin}{\passmax} \\
      9  & \heatcell{33}{\passmin}{\passmax} & \heatcell{27}{\passmin}{\passmax} & \heatcell{20}{\passmin}{\passmax} & \heatcell{14}{\passmin}{\passmax} & \heatcell{9}{\passmin}{\passmax}  \\
      10 & \heatcell{28}{\passmin}{\passmax} & \heatcell{23}{\passmin}{\passmax} & \heatcell{16}{\passmin}{\passmax} & \heatcell{11}{\passmin}{\passmax} & \heatcell{7}{\passmin}{\passmax}  \\
      \bottomrule
    \end{tabular}
    \par\smallskip
    {\small (b) $\Qeasy$ and $\Qhard$ (6--10 terms \& digits)}
  \end{minipage}
\end{table}

%% file: figures/task_example.tex

\begin{taskexampletwo}{Example: \texttt{chain sum} (4 terms, 3 digits)}
  \textbf{Prompt:} \textit{Compute the following step by step: $381 - 706 - 602 + 383 =$ Show your reasoning and return the final answer in} \texttt{\char92 boxed\char123\char125} \textit{tags, for example} \texttt{\char92 boxed\char123 42\char125}\textit{.}

  \tcblower

  \begin{tabular}{@{}p{0.47\linewidth}@{\hspace{0.6em}}|@{\hspace{0.6em}}p{0.47\linewidth}@{}}
    \textbf{Completion A:} To solve this problem, we need to follow the order of operations (PEMDAS/BODMAS) which means we do the operations from left to right since there are no parentheses, exponents, multiplication/division, or addition/subtraction operations to perform.\newline
    \newline
    Let's start with the first two numbers:\newline
    381 - 706 = -325\newline
    \newline
    Now, we subtract the third number from the result:\newline
    -325 - 602 = -927\newline
    \newline
    Finally, we add the last number:\newline
    -927 + 383 = -544\newline
    \newline
    So, the final answer is:\newline
    \boxed{-544}
    &
    \textbf{Completion B:} To compute the expression 381 - 706 - 602 + 383, we can follow these steps:\newline
    \newline
    1. Subtract 706 from 381:\newline
    381 - 706 = -325\newline
    \newline
    2. Subtract 602 from -325:\newline
    -325 - 602 = -927\newline
    \newline
    3. Add 383 to -927:\newline
    -927 + 383 = -544\newline
    \newline
    So, the final answer is \boxed{-544}.
  \end{tabular}
\end{taskexampletwo}

%% file: sections/hyperparameters.tex
\section{Hyperparameters}\label{app:hyperparameters}
We list all training, optimization, and method-specific hyperparameters used throughout our experiments. Our comparison is \emph{data-constrained}: every method trains on the same data pool, so differences in downstream performance reflect how effectively each objective extracts signal from a fixed dataset rather than differences in data quantity or coverage.
We train each method with LoRA adapters using AdamW and a cosine learning-rate schedule. For each configuration, we train for 1--4 epochs and select the checkpoint with the highest $\text{pass}@1$ on a held-out validation set. Default hyperparameters follow established conventions for each method~\citep{huLoRALowRankAdaptation2021, rafailovDirectPreferenceOptimization2024, yuanRRHFRankResponses2023}; we ablate key choices ($\lambda$ for RRHF, pair selection for DPO) in \cref{app:rank_weight,app:results-randomvsjudge}. Hyperparameters and method-specific additions are listed in \cref{tab:hyperparams-unified}; values shown as sets indicate the values swept, and we report results for the best-performing configuration per method.

\subsection{Evaluation Configuration}
\cref{tab:hyperparams-eval} summarizes the evaluation setup used for all reported results. All reported $\text{pass}@1$ and $\text{pass}@8$ values are given in \%, estimated using the unbiased estimator of \citet{chenEvaluatingLargeLanguage2021}. With $n=8$ samples per problem, $\text{pass}@8$ reduces to a binary metric: whether at least one sample solves the problem.
For both \texttt{chain sum} and GSM8K, we check for answer equivalence using Math-Verify~\citep{kydlicekMathVerifyMathVerification2026}.
\begin{table}[tb]
  \caption{Evaluation configuration.}
  \label{tab:hyperparams-eval}
  \centering
  \begin{tabularx}{0.7\linewidth}{lX}
    \toprule
    Parameter & Value \\
    \midrule
	    Evaluation dataset & $\texttt{hannoh/chainsum\_eval}$ \newline $GSM8K$ (test split) \\
	    Number of questions & $200$ (\texttt{chain sum}) \newline $1{,}319$ (GSM8K) \\
    Samples per question ($n$) & 8 \\
    Temperature & 0.6 \\
    Top-$p$ & 0.95 \\
    Max generation tokens & 2048 \\
    \bottomrule
  \end{tabularx}
\end{table}

\subsection{Judge Configuration}
\cref{tab:hyperparams-judge} lists the configuration of the LLM-as-a-judge used to score completions during training data collection.

\paragraph{Few-shot prompting for judging} For base models, we employ few-shot prompting with worked examples to reliably parse judge outputs. For instruction-tuned models, we use standard zero-shot prompting with evaluation criteria. Prompts are adapted per task (e.g.,\ \texttt{chain sum} vs.\ GSM8K); full prompts are in \cref{app:prompts}.

\begin{table}[tb]
  \caption{LLM-as-a-judge configuration.}
  \label{tab:hyperparams-judge}
  \centering
  \begin{tabularx}{0.7\linewidth}{lX}
    \toprule
    Parameter & Value \\
    \midrule
    Judge model & $J_\theta$: $\texttt{Qwen/Qwen3-1.7B-Base}$ \newline $J_\phi$: $\texttt{Qwen/Qwen3-30B}$ \\
    Tournament group size & 2 (pairwise) \\
    Temperature & 0 (greedy) \\
    Output constraint & Regex: ends with \texttt{Judgment: [N]} \\
    Max generation tokens & 1024 \\
    \bottomrule
  \end{tabularx}
\end{table}

We use pairwise comparisons (group size 2) for the judge. Larger group sizes would allow ranking more completions in a single pass, but are likely to degrade judgment quality when using a small self-judge model (especially for base models) and additionally require proportionally larger context windows to accommodate all completions simultaneously.

The output is constrained by a regex that requires the response to end with \texttt{Judgment: [N]}. Because we use a base model as the judge rather than an instruction-tuned model, it does not reliably follow formatting instructions from the prompt alone. The regex constraint ensures that every judge response produces a parseable selection, avoiding the need to discard malformed outputs.

\subsection{Training Hyperparameters}
\cref{tab:hyperparams-unified} lists the hyperparameters shared across all training runs, along with method-specific additions.

We train for 1--4 epochs and select the best checkpoint by $\text{pass}@1$ on the held-out validation set.
We set the LoRA alpha to twice the rank ($\alpha = 2r$), following the practical recommendation of~\citet{phdPracticalTipsFinetuning2025}.
For DPO we disable weight decay, following~\citet{ivisonUnpackingDPOPPO2024}.
The DPO $\beta$ and RRHF rank loss weight $\lambda$ were selected by hyperparameter sweeps described in \cref{app:dpo-hyperparam,app:rank_weight}, respectively.
Effective batch sizes are achieved via gradient accumulation across per-device batches; how the batch is divided between accumulation steps and per-device batch size did not affect results, except for the ranking loss in $\sigma$-RRHF where the set of completions compared within each step depends on the per-device batch size, which we ensure is always at least $n$ so that all completions for a question appear together in the same batch.

\begin{table}[tb]
  \caption{Training hyperparameters. Sets indicate the values swept; we report results for the best-performing configuration per method.}
  \label{tab:hyperparams-unified}
  \centering
  \begin{tabularx}{0.75\linewidth}{llX}
    \toprule
    Symbol & Value & Description \\
    \midrule
    \multicolumn{3}{c}{Training}\\
    \midrule
    $M_{base}$ & \texttt{Qwen3-1.7B-Base} & Base model. \\
    $E$ & $\{1, 2, 4\}$ & Training epochs. \\
    $B$ & $\{2, 4, 8, 16\}$ & Batch size. \\
    $L$ & 2048 & Max sequence length. \\
    \midrule
    \multicolumn{3}{c}{Optimization (AdamW)}\\
    \midrule
    $\eta$ & $\{5 \times 10^{-6}, 10^{-5}, 5 \times 10^{-5}\}$ & Learning rate. \\
    Scheduler & Cosine & Cosine LR decay. \\
    Warmup & 0.1 & Linear warmup. \\
    $w_d$ & 0.01 & Weight decay (except for DPO). \\
    \midrule
    \multicolumn{3}{c}{LoRA}\\
    \midrule
    $r$ & 32 & LoRA rank. \\
    $\alpha$ & 64 & LoRA alpha. \\
    $p_{\text{drop}}$ & 0.05 & LoRA dropout. \\
    \midrule
    \multicolumn{3}{c}{Scoring}\\
    \midrule
    $\epsilon$ & $10^{-8}$ & Floor for $\bar H$ in inverse-entropy scoring (\cref{sec:scoring-methods}). \\
    \midrule
    \multicolumn{3}{c}{DPO}\\
    \midrule
    $\beta$ & $\{0.1, 0.3, 0.5, 2, 5\}$ & KL penalty coefficient. \\
    $w_d$ (DPO) & 0.0 & No weight decay. \\
    Loss & Sigmoid & Loss function type. \\
    \midrule
    \multicolumn{3}{c}{RRHF}\\
    \midrule
    $\lambda$ & $\{0, 0.01, 0.1, 0.5, 1.0\}$ & Rank loss weight. \\
    $\sigma$ & Logistic weighting & Scales each hinge-loss pair by $\sigma(r_j - r_i)$, the logistic sigmoid of the score gap (\cref{sec:rrhf}). \\
    \bottomrule
  \end{tabularx}
\end{table}

%% file: figures/results_hard.tex

\pgfplotstableread[col sep=comma]{data/chainsum_hard_results.csv}\resultshard
\begin{table}[tb]
  \caption{
    \textbf{Method comparison} on $\Qhard$ chain sum problems.
    \mbox{+ DPO (correctness)} uses correct-vs-incorrect pairs from $\Qhard$ (\cref{sec:methods-dpo}).
    \textbf{Bold} is best per column.
  }
  \label{tab:results-hard}
  \centering
  \setlength{\tabcolsep}{4pt}
  \pgfplotstabletypeset[
    outfile=figures/pgf-static/results_hard.tex,
    begin table=\begin{tabularx}{\columnwidth},
    end table=\end{tabularx},
    col sep=comma,
    string type,
    columns={method, data, pass1, pass8},
    columns/method/.style={column name=Model / Training, column type=l},
    columns/data/.style={column name=Scoring, column type=X},
    columns/pass1/.style={column name=pass@1, column type=r},
    columns/pass8/.style={column name=pass@8, column type=r},
    every head row/.style={before row=\toprule, after row=\midrule},
    every last row/.style={after row=\bottomrule},
    every row no 1/.style={before row={\arrayrulecolor{black!20}\midrule\arrayrulecolor{black}}},
    every row no 3/.style={before row={\arrayrulecolor{black!20}\midrule\arrayrulecolor{black}}},
    every row 2 column 2/.style={postproc cell content/.append style={/pgfplots/table/@cell content/.add={\bfseries}{}}},
    every row 2 column 3/.style={postproc cell content/.append style={/pgfplots/table/@cell content/.add={\bfseries}{}}},
  ]{\resultshard}
\end{table}

%% file: figures/dpo_reward_margins.tex

\pgfplotstableread[col sep=comma]{data/chainsum_dpo_hard.csv}\dpoRandHardM
\pgfplotstableread[col sep=comma]{data/chainsum_dpo_best_contrastive_hard.csv}\dpoJudgeHardM

\begin{figure}[tb]
  \centering

  \pgfplotsset{
    dpoaxis/.style={
      width=0.94\linewidth,
      height=5cm,
      xlabel={Training progress},
      xmin=0, xmax=1,
      ymin=0,
      legend style={font=\scriptsize, draw=none, fill=none, at={(0.98,0.02)}, anchor=south east},
      axis background/.style={fill=gray!4},
      axis x line=bottom,
      axis y line=left,
      axis line style={-},
      tick align=outside,
      xtick={0,0.25,0.5,0.75,1.0},
      xticklabels={0,,.5,,1},
    },
    raw/.style={thin, opacity=0.2},
    avg/.style={thick},
  }

  \begin{subfigure}[t]{0.48\linewidth}
    \centering
    \tikzfigname{dpo_reward_margins_margin}
    \begin{tikzpicture}
      \begin{axis}[
        dpoaxis,
        ylabel={Reward margin (\%)},
        ymax=21,
        ytick={0,5,10,15,20},
        yticklabels={0\%,5\%,10\%,15\%,20\%},
      ]
        \addplot[raw, color=plotblue, forget plot]   table[x=step_norm, y={train/rewards/margins}]{\dpoRandHardM};
        \addplot[avg, color=plotblue]   table[x=step_norm, y={train/rewards/margins_avg}]{\dpoRandHardM};
        \addlegendentry{DPO (random)}

        \addplot[raw, color=plotorange, forget plot]  table[x=step_norm, y={train/rewards/margins}]{\dpoJudgeHardM};
        \addplot[avg, color=plotorange]  table[x=step_norm, y={train/rewards/margins_avg}]{\dpoJudgeHardM};
        \addlegendentry{DPO (judge)}
      \end{axis}
    \end{tikzpicture}
    \caption{Implicit reward margin $\hat{r}(x, y_w) - \hat{r}(x, y_l)$ over training.}
    \label{fig:dpo-reward-margins-margin}
  \end{subfigure}
  \hfill
  \begin{subfigure}[t]{0.48\linewidth}
    \centering
    \tikzfigname{dpo_reward_margins_eval}
    \begin{tikzpicture}
      \begin{axis}[
        dpoaxis,
        ylabel={pass@1 (\%)},
        xlabel={Training progress},
        xmin=0, xmax=1,
        ymin=0.08, ymax=0.38,
        xtick={0.25,0.5,0.75,1.0},
        xticklabels={,.5,,1},
        ytick={0.10,0.15,0.20,0.25,0.30,0.35},
        yticklabels={10\%,15\%,20\%,25\%,30\%,35\%},
      ]
        \addplot[avg, color=plotblue, mark=*, mark size=1.5pt] coordinates {
          (0,  0.1063)
          (0.25,  0.1850)
          (0.5, 0.2981)
          (0.75, 0.2963)
          (1.0, 0.2956)
        };
        \addlegendentry{DPO (random)}

        \addplot[avg, color=plotorange, mark=square*, mark size=1.5pt] coordinates {
          (0,  0.1063)
          (0.25,  0.1856)
          (0.5, 0.2963)
          (0.75, 0.3494)
          (1.0, 0.3388)
        };
        \addlegendentry{DPO (judge)}
      \end{axis}
    \end{tikzpicture}
    \caption{pass@1 at four evaluation checkpoints.}
    \label{fig:dpo-reward-margins-eval}
  \end{subfigure}

  \caption{%
    DPO training dynamics on $\Qhard$ for random vs.\ judge-selected pair construction.%
  }
  \label{fig:dpo-reward-margins}
\end{figure}

%% file: figures/rank_weight_eval.tex

\pgfplotstableread[col sep=comma]{data/rrhf_rank_weight_ablation.csv}\rankweight

\definecolor{cRWpass1}{HTML}{2C7BB6}
\definecolor{cRWpass8}{HTML}{DD8452}

\begin{figure}[tb]
  \centering
  \begin{subfigure}[t]{0.48\linewidth}
    \centering
    \tikzfigname{rank_weight_pass1}
    \begin{tikzpicture}
      \begin{axis}[
        width=\linewidth,
        height=5cm,
        xmode=log,
        xmin=0.0015, xmax=2.5,
        xtick={0.003, 0.01, 0.1, 0.5, 1.0},
        xticklabels={$0$, $10^{-2}$, $10^{-1}$, $0.5$, $1$},
        xlabel={Rank loss weight $\lambda$},
        ylabel={Accuracy (\%)},
        ymin=33, ymax=43,
        ytick={34,36,38,40,42},
        axis background/.style={fill=gray!4},
        axis x line=bottom,
        axis y line=left,
        axis line style={-},
        tick align=outside,
        mark size=2.5pt,
        ]
        \addplot[
          color=cRWpass1, thick, solid,
          mark=*, mark options={fill=cRWpass1},
          error bars/.cd,
          y dir=both, y explicit,
          error bar style={thick},
          ] table[
          x expr=\thisrow{lambda}<0.001 ? 0.003 : \thisrow{lambda},
          y=pass1_mean,
          y error=pass1_std,
          ] {\rankweight};
      \end{axis}
    \end{tikzpicture}
    \caption{pass@1}
    \label{fig:rank-weight-pass1}
  \end{subfigure}
  \hfill
  \begin{subfigure}[t]{0.48\linewidth}
    \centering
    \tikzfigname{rank_weight_pass8}
    \begin{tikzpicture}
      \begin{axis}[
        width=\linewidth,
        height=5cm,
        xmode=log,
        xmin=0.0015, xmax=2.5,
        xtick={0.003, 0.01, 0.1, 0.5, 1.0},
        xticklabels={$0$, $10^{-2}$, $10^{-1}$, $0.5$, $1$},
        xlabel={Rank loss weight $\lambda$},
        ylabel={Accuracy (\%)},
        ymin=70, ymax=80,
        ytick={71,73,75,77,79},
        axis background/.style={fill=gray!4},
        axis x line=bottom,
        axis y line=left,
        axis line style={-},
        tick align=outside,
        mark size=2.5pt,
        ]
        \addplot[
          color=cRWpass8, thick, solid,
          mark=square*, mark options={fill=cRWpass8},
          error bars/.cd,
          y dir=both, y explicit,
          error bar style={thick},
          ] table[
          x expr=\thisrow{lambda}<0.001 ? 0.003 : \thisrow{lambda},
          y=pass8_mean,
          y error=pass8_std,
          ] {\rankweight};
      \end{axis}
    \end{tikzpicture}
    \caption{pass@8}
    \label{fig:rank-weight-pass8}
  \end{subfigure}
  \caption{%
    Rank loss weight ablation on $\Qhard$ chain sum problems. The x-axis is log-scaled; $\lambda\!=\!0$ (no rank loss) is placed at the left as a reference point. Pass@1 peaks at $\lambda\!=\!10^{-2}$ and degrades monotonically for larger weights. Pass@8 peaks at $\lambda\!=\!10^{-1}$ with no clear monotonic trend, but some rank loss has an effect on generalization to OOD questions. Responses were ranked with the Self-judge.%
  }
  \label{fig:rank-weight-eval}
\end{figure}

%% file: figures/rrhf_variants.tex

\pgfplotstableread[col sep=comma]{data/rrhf_variant_ablation.csv}\rrhfvariants

\begin{table}[tb]
  \caption{
    $\sigma$-RRHF ablation on $\Qhard$ chain sum.
    The SFT anchoring term is essential: without it performance collapses to near-random.
    Hinge loss and logistic reweighting provide no consistent improvement.
  }
  \label{tab:rrhf-variants}
  \centering
  \pgfplotstabletypeset[
    outfile=figures/pgf-static/rrhf_variants.tex,
    col sep=comma,
    string type,
    columns={variant, pass1, pass8},
    columns/variant/.style={column name=Variant, column type=l},
    columns/pass1/.style={column name=pass@1, column type=r},
    columns/pass8/.style={column name=pass@8, column type=r},
    every head row/.style={before row=\toprule, after row=\midrule},
    every last row/.style={after row=\bottomrule},
  ]{\rrhfvariants}
\end{table}

%% file: figures/dpo_strict_sweep_eval.tex

\pgfplotstableread[col sep=comma]{data/dpo_beta_sweep.csv}\dpohypertable
\begin{table}[tb]
  \caption{%
    DPO hyperparameter sweep on the $\Qstrict$ split (200 questions, 8 samples per question), using self-judge pair selection.
    \textbf{Bold} indicates the best configuration by pass@1.%
  }
  \label{tab:dpo-strict-sweep}
  \centering
  \pgfplotstabletypeset[
    outfile=figures/pgf-static/dpo_strict_sweep_eval.tex,
    col sep=comma,
    string type,
    columns={beta, lr, epochs, pass1, pass8},
    columns/beta/.style={  column name=$\beta$,      column type=r},
    columns/lr/.style={    column name=$\eta$,       column type=r},
    columns/epochs/.style={column name=Epochs,       column type=r},
    columns/pass1/.style={ column name=pass@1,        column type=r},
    columns/pass8/.style={ column name=pass@8,        column type=r},
    every head row/.style={before row=\toprule, after row=\midrule},
    every last row/.style={after row=\bottomrule},
    every row 4 column 0/.style={postproc cell content/.append style={/pgfplots/table/@cell content/.add={\bfseries}{}}},
    every row 4 column 1/.style={postproc cell content/.append style={/pgfplots/table/@cell content/.add={\bfseries}{}}},
    every row 4 column 2/.style={postproc cell content/.append style={/pgfplots/table/@cell content/.add={\bfseries}{}}},
    every row 4 column 3/.style={postproc cell content/.append style={/pgfplots/table/@cell content/.add={\bfseries}{}}},
    every row 4 column 4/.style={postproc cell content/.append style={/pgfplots/table/@cell content/.add={\bfseries}{}}},
  ]{\dpohypertable}
\end{table}

%% file: figures/gsm8k_dpo_training.tex

\pgfplotstableread[col sep=comma]{data/gsm8k_dpo_entropy.csv}\gsmDpoEntropy
\pgfplotstableread[col sep=comma]{data/gsm8k_dpo_sjudge.csv}\gsmDpoSjudge
\pgfplotstableread[col sep=comma]{data/gsm8k_dpo_strongjudge.csv}\gsmDpoStrongjudge

\begin{figure}[tb]
  \definecolor{cEntropy}{HTML}{4C72B0}
  \definecolor{cSjudge}{HTML}{DD8452}
  \definecolor{cStrong}{HTML}{55A868}

  \pgfplotsset{
    raw/.style={thin, opacity=0.2},
    avg/.style={thick},
  }
  \centering
  \tikzfigname{gsm8k_dpo_training}
  \begin{tikzpicture}
    \begin{axis}[
      width=0.65\linewidth,
      height=5cm,
      xlabel={Training progress},
      ylabel={Reward margin},
      xmin=0, xmax=1,
      ymin=0,
      axis background/.style={fill=gray!4},
      axis x line=bottom,
      axis y line=left,
      axis line style={-},
      tick align=outside,
      xtick={0,0.25,0.5,0.75,1.0},
      xticklabels={0,,.5,,1},
      ]
      \addplot[raw, color=cEntropy, forget plot]
      table[x=step_norm, y={train/rewards/margins}]{\gsmDpoEntropy};
      \addplot[avg, color=cEntropy]
      table[x=step_norm, y={train/rewards/margins_avg}]{\gsmDpoEntropy};

      \addplot[raw, color=cStrong, forget plot]
      table[x=step_norm, y={train/rewards/margins}]{\gsmDpoStrongjudge};
      \addplot[avg, color=cStrong]
      table[x=step_norm, y={train/rewards/margins_avg}]{\gsmDpoStrongjudge};

      \addplot[raw, color=cSjudge, forget plot]
      table[x=step_norm, y={train/rewards/margins}]{\gsmDpoSjudge};
      \addplot[avg, color=cSjudge]
      table[x=step_norm, y={train/rewards/margins_avg}]{\gsmDpoSjudge};
    \end{axis}
  \end{tikzpicture}
  \caption{%
    Implicit reward margin $\hat{r}(x, y_w) - \hat{r}(x, y_l)$ over training on GSM8K.
    \textcolor{cEntropy}{Inverse entropy} reaches the highest margin;
    \textcolor{cStrong}{Qwen3-30B} grows more steadily;
    \textcolor{cSjudge}{Qwen3-1.7B-Base} (self-judge) stays near zero.
    Faint lines are raw per-step values; bold lines are rolling averages.
  }
  \label{fig:gsm8k-dpo-training}
\end{figure}

%% file: figures/table_judge_agreement.tex
\begin{table*}[tb]
  \centering
  \small
  \caption{Within-question Spearman rank correlations between scoring methods ($r_i^{J_\phi}$: strong judge, $r_i^{J_\theta}$: self-judge, $r_i^{H}$: inverse entropy) on $\Qstrict$ \texttt{chain sum} and $\Qeasy$ GSM8K. Each cell reports mean (median) $\rho$ and the percentage of questions with $\rho > 0$.}
  \label{tab:judge-agreement}
  \resizebox{\textwidth}{!}{\begin{tabular}{lcccc}
    \toprule
    & \multicolumn{2}{c}{\texttt{chain sum} ($\Qstrict$)} & \multicolumn{2}{c}{GSM8K ($\Qeasy$)} \\
    \cmidrule(lr){2-3} \cmidrule(lr){4-5}
    Pair & $\bar{\rho}$ ($\tilde{\rho}$) & \% pos.\ & $\bar{\rho}$ ($\tilde{\rho}$) & \% pos.\ \\
    \midrule
    $r_i^{J_\phi}$ vs.\ $r_i^{J_\theta}$ (\texttt{Qwen3-30B} vs.\ \texttt{Qwen3-1.7B-Base}) & $+$0.15 ($+$0.19) & 62.0 & $-$0.20 ($-$0.24) & 27.9 \\
    $r_i^{J_\phi}$ vs.\ $r_i^{H}$ (\texttt{Qwen3-30B} vs.\ Inverse entropy) & $+$0.44 ($+$0.50) & 89.4 & $+$0.56 ($+$0.63) & 94.0 \\
    $r_i^{J_\theta}$ vs.\ $r_i^{H}$ (\texttt{Qwen3-1.7B-Base} vs.\ Inverse entropy) & $+$0.16 ($+$0.19) & 68.5 & $-$0.20 ($-$0.22) & 30.9 \\
    \bottomrule
  \end{tabular}}
\end{table*}

%% file: sections/prompts.tex
\section{Prompts}\label{app:prompts}

The following prompts are used for LLM-as-a-judge completion ranking
(\cref{sec:scoring-methods}).
Each task uses two prompt formulations: a \emph{full few-shot} version that
embeds a worked example, and a \emph{short system\,/\,user} pair without a
demonstration.
The judge generates free-form analysis constrained by a regex to end with
\texttt{Judgment:~[IDX]}.


\subsection{Prompts for \texttt{chain sum}}\label{app:prompts-chainsum}

\subsubsection{Select Best Correct Completion}\label{app:prompt-chainsum-best}

Identifies the highest-quality correct completion among multiple correct solutions.
The judge evaluates step-by-step correctness, logical progression, notation,
and efficiency.

\medskip\noindent\emph{Full few-shot prompt (used when \texttt{instruct=False}):}

\ifverboseprompts
\begin{prompt}{\texttt{chain sum} -- Select Best Correct Completion (Full Few-Shot)}
  # Mathematical Solution Evaluation

  There are math problems followed by a number of solutions. The task is to
  systematically analyze these solutions to identify the most mathematically
  sound approach.

  ### Evaluation Process

  #### 1. Initial Screening
  * Group solutions by their final answers.
  * Identify and explain mathematical contradictions between different answers.
  * Eliminate solutions with clear mathematical errors.

  #### 2. Detailed Analysis
  For remaining solutions, evaluate:
  * **Mathematical Precision and Accuracy:** Correctness of calculations and
  theorem applications.
  * **Logical Progression:** The flow and sequence of steps.
  * **Completeness:** Whether the mathematical reasoning is fully documented.
  * **Notation:** Proper use of mathematical symbols and the `\boxed{}` format.
  * **Special Conditions:** Handling of edge cases or specific constraints.
  * **Error Correction:** For solutions containing and addressing errors,
  evaluate the identification and correction methodology.

  #### 3. Solution Comparison
  Compare viable solutions based on:
  * **Efficiency:** The elegance of the approach.
  * **Clarity:** How well the reasoning is communicated.
  * **Sophistication:** The depth of the method used.
  * **Robustness:** Whether the solution works for all cases.

  ### Response Requirements
  1. **Brief Analysis** of conflicting answers.
  2. **Detailed Evaluation** of mathematically sound solutions.
  3. **Justification** for eliminating incorrect solutions.
  4. **Clear Explanation** for selecting the best approach.

  An evaluation ends with:
  **Judgment: [IDX]** *(Where `IDX` is the index $0 - {max_idx}$ of the best
  solution.)*

  ---

  ## EVALUATION A
  **Problem:** Compute the following step by step:
  593615 + 204846 - 838944 + 816336 - 913166 - 338746 =

  **Solutions (all correct, ground truth = -476059):**
0) Let's solve the problem step by step:

1. **Addition:**
\\[
  593615 + 204846 = 798461
\\]

2. **Subtraction:**
\\[
  798461 - 838944 = -40483
\\]

3. **Addition:**
\\[
  -40483 + 816336 = 775853
\\]

4. **Subtraction:**
\\[
  775853 - 913166 = -137313
\\]

5. **Subtraction:**
\\[
  -137313 - 338746 = -476059
\\]

The final answer is: \\boxed{-476059}

1) Let's solve this step by step:

1. Compute 593615 + 204846
2. Subtract 838944 from the result
3. Add 816336 to the new result
4. Subtract 913166 from the new result
5. Subtract 338746 from the new result

Now, let's compute each step:

1. 593615 + 204846 = 798461
2. 798461 - 838944 = -40483
3. -40483 + 816336 = 775853
4. 775853 - 913166 = -137313
5. -137313 - 338746 = -476059

So, the final answer is \\boxed{-476059}.

### EVALUATION
**Initial Screening:** Both solutions arrive at -476059. No answer
disagreement; both are arithmetically correct.

**Analysis:**
- Solution 0: Proceeds directly to computation. Each step is labeled by
operation type (**Addition:**, **Subtraction:**) and formatted in LaTeX
display math. The structure is clean: one numbered item per operation,
intermediate result immediately visible.
- Solution 1: Opens with a planning phase that lists all five steps as
instructions before computing anything. The plan is redundant given the
computations that follow. The inline `number = result` style lacks the
visual clarity of display math.

**Comparison:** Both are correct and complete. Solution 0 is more direct --
it computes each step immediately without a preamble. The operation-type
labels and LaTeX formatting make intermediate values easy to scan and verify.
Solution 1's two-phase structure (plan then execute) adds length without
adding information.

**Judgment: [0]**

---

## EVALUATION B
**Problem:** {problem}

**Solutions:** {solutions}

### EVALUATION
\end{prompt}
\else
\begin{prompt}{\texttt{chain sum} -- Select Best Correct Completion (Full Few-Shot, condensed)}
# Mathematical Solution Evaluation

[Evaluation criteria: Initial Screening (group by answer, eliminate errors),
Detailed Analysis (precision, logical progression, completeness, notation,
special conditions), Solution Comparison (efficiency, clarity, sophistication,
robustness).]

Response ends with: **Judgment: [IDX]**  (IDX in 0-{max_idx})

---

## EVALUATION A
**Problem:** Compute: 593615 + 204846 - 838944 + 816336 - 913166 - 338746 =
**Solutions (ground truth = -476059):**
0) [Computes each step directly with display math and operation-type labels.]
Final: \boxed{-476059}
1) [Lists all steps first as a plan, then computes inline.  \boxed{-476059}]

### EVALUATION
[Both correct. Solution 0 preferred: direct, labeled, display math.
Solution 1 has redundant planning phase.]  **Judgment: [0]**

---

## EVALUATION B
**Problem:** {problem}   **Solutions:** {solutions}   ### EVALUATION
\end{prompt}
\fi

\subsection{GSM8K Prompts}\label{app:prompts-gsm8k}

\subsubsection{Select Best Correct Completion}\label{app:prompt-gsm8k-best}

Same role as the \texttt{chain sum} variant but adapted for grade-school math word
problems: the evaluation criteria add problem comprehension, unit-conversion
accuracy, and logical flow alongside the structural quality criteria.

\medskip\noindent\emph{Full few-shot prompt (used when \texttt{instruct=False}):}

\ifverboseprompts
\begin{prompt}{GSM8K -- Select Best Correct Completion (Full Few-Shot)}
# Word Problem Solution Evaluation

There are word problems followed by a number of solutions. The task is to systematically analyze these solutions to identify the most mathematically sound approach.

### Evaluation Process

#### 1. Initial Screening
* Group solutions by their final answers.
* Identify and explain contradictions between different answers.
* Eliminate solutions with clear mathematical or logical errors.

#### 2. Detailed Analysis
For remaining solutions, evaluate:
* **Problem Comprehension:** Does the solution correctly identify all given quantities and what is being asked?
* **Step-by-step Correctness:** Are all arithmetic operations and unit conversions accurate?
* **Logical Flow:** Does each step follow naturally from the previous one?
* **Completeness:** Are all sub-problems addressed before combining into a final answer?
* **Clarity:** Is the solution easy to follow without unnecessary repetition?

#### 3. Solution Comparison
Compare viable solutions based on:
* **Directness:** Does the solution get to the point without redundant preambles?
* **Organization:** Are the steps well-structured and labeled?
* **Presentation:** Is the final answer clearly stated?

### Response Requirements
1. **Brief Analysis** of any answer disagreements.
2. **Detailed Evaluation** of mathematically sound solutions.
3. **Clear Explanation** for selecting the best approach.

An evaluation ends with:
**Judgment: [IDX]** *(Where `IDX` is the index $0 - {max_idx}$ of the best solution.)*

---

## EVALUATION A
**Problem:** For every 12 cans you recycle, you receive $0.50, and for every 5 kilograms of newspapers, you receive $1.50. If your family collected 144 cans and 20 kilograms of newspapers, how much money would you receive?

**Solutions (all correct, ground truth = 12):**
0) To solve this problem, we need to calculate the amount of money received from recycling the cans and the newspapers separately, and then add those amounts together.

1. **Calculate the money received from recycling cans:**

   We know that for every 12 cans, you receive $0.50. Your family collected 144 cans. To find out how many sets of 12 cans they have, we divide the total number of cans by 12:
   \\\\[
     \\\\frac{{144 \\\\text{{ cans}}}}{{12 \\\\text{{ cans/set}}}} = 12 \\\\text{{ sets}}
   \\\\]

   Since each set of 12 cans gives $0.50, we multiply the number of sets by $0.50:
   \\\\[
     12 \\\\text{{ sets}} \\\\times \\\\$0.50/\\\\text{{set}} = \\\\$6.00
   \\\\]

2. **Calculate the money received from recycling newspapers:**

   We know that for every 5 kilograms of newspapers, you receive $1.50. Your family collected 20 kilograms. To find out how many sets of 5 kilograms they have:
   \\\\[
     \\\\frac{{20 \\\\text{{ kg}}}}{{5 \\\\text{{ kg/set}}}} = 4 \\\\text{{ sets}}
   \\\\]

   Since each set of 5 kilograms gives $1.50:
   \\\\[
     4 \\\\text{{ sets}} \\\\times \\\\$1.50/\\\\text{{set}} = \\\\$6.00
   \\\\]

3. **Add the amounts:**
   \\\\[
     \\\\$6.00 + \\\\$6.00 = \\\\$12.00
   \\\\]

Therefore, the total amount of money your family would receive is $\\\\boxed{{12}}$.

1) Let's solve this step by step:

1. Calculate the money earned from recycling cans:
   - For every 12 cans, you receive $0.50.
   - Your family collected 144 cans.
   - To find out how many sets of 12 cans there are in 144 cans, divide 144 by 12: 144 / 12 = 12 sets.
   - Multiply the number of sets by the amount earned per set: 12 sets * $0.50/set = $6.00.

2. Calculate the money earned from recycling newspapers:
   - For every 5 kilograms of newspapers, you receive $1.50.
   - Your family collected 20 kilograms of newspapers.
   - To find out how many sets of 5 kilograms there are in 20 kilograms, divide 20 by 5: 20 / 5 = 4 sets.
   - Multiply the number of sets by the amount earned per set: 4 sets * $1.50/set = $6.00.

3. Add the money earned from both cans and newspapers:
   - Money earned from cans: $6.00
   - Money earned from newspapers: $6.00
   - Total money earned: $6.00 + $6.00 = $12.00

Final answer: You would receive $12.00.

### EVALUATION
**Initial Screening:** Both solutions arrive at $12.00. No answer disagreement; both are arithmetically correct.

**Analysis:**
- Solution 0: Each sub-problem is clearly labeled (**Calculate the money received from recycling cans/newspapers**). Uses LaTeX display math for key calculations, making intermediate values easy to verify at a glance. The structure is clean with numbered sections and a boxed final answer.
- Solution 1: Correct and complete, but uses a flat inline style throughout. The final answer lacks \\\\boxed formatting. A separate summary block ("Money earned from cans: $6.00...") restates values already computed, adding length without information.

**Comparison:** Both are correct. Solution 0 is better organized -- bold section headers, LaTeX for key steps, and a boxed answer make it easier to scan and verify. Solution 1 is slightly redundant with its summary block.

**Judgment: [0]**

---

## EVALUATION B
**Problem:** A baker makes 48 cookies. He keeps one-quarter for himself and divides the rest equally among 3 friends. How many cookies does each friend get?

**Solutions (all correct, ground truth = 12):**
0) The total number of cookies baked is 48.

Step 1: The baker keeps one-quarter for himself.
One-quarter of 48: 48 / 4 = 12 cookies kept.

Step 2: Find the remaining cookies.
Remaining = 48 - 12 = 36 cookies.

Step 3: Divide among 3 friends.
Each friend gets: 36 / 3 = 12 cookies.

Check: 12 + 12 + 12 + 12 = 48 (ok)

Each friend receives $\\\\boxed{{12}}$ cookies.

1) Cookies the baker keeps: 48 / 4 = 12

Remaining to share: 48 - 12 = 36

Each friend gets: 36 / 3 = 12

$\\\\boxed{{12}}$

### EVALUATION
**Initial Screening:** Both solutions reach 12. No disagreement.

**Analysis:**
- Solution 0: Correct and systematic, but adds a verification step (12 + 12 + 12 + 12 = 48) that restates values already established by the arithmetic. The labeled steps are helpful, but the check adds length without new information.
- Solution 1: Three concise lines, one arithmetic step each. Reaches the same conclusion without redundancy.

**Comparison:** Both are correct. Solution 1 is more direct -- the verification in Solution 0 is unnecessary when every step is already clearly shown. Solution 1 is preferred.

**Judgment: [1]**

---

## EVALUATION C
**Problem:** A car drives at 60 mph for 2 hours, then at 40 mph for 1.5 hours. How many miles does it travel in total?

**Solutions (all correct, ground truth = 180):**
0) The car drives at 60 mph for 2 hours, so that's 60 * 2 = 120 miles. Then it goes at 40 mph for 1.5 hours, which is 40 * 1.5 = 60 miles. Total: 120 + 60 = 180 miles. The answer is $\\\\boxed{{180}}$.

1) Distance in first segment: 60 mph * 2 hours = 120 miles

Distance in second segment: 40 mph * 1.5 hours = 60 miles

Total distance: 120 + 60 = 180 miles

$\\\\boxed{{180}}$

### EVALUATION
**Initial Screening:** Both solutions reach 180. No disagreement.

**Analysis:**
- Solution 0: Correct, but merges all steps into a continuous sentence. While compact, it makes it harder to scan each individual computation.
- Solution 1: Computes each segment on a dedicated labeled line; the final sum is on its own line. Each step is easy to locate and verify in isolation.

**Comparison:** Solution 1's one-operation-per-line structure is easier to read and audit than Solution 0's run-on prose. Solution 1 is preferred for clarity.

**Judgment: [1]**

---

## EVALUATION D
**Problem:** {problem}

**Solutions (indices 0--{max_idx}):**
{solutions}

### EVALUATION
\end{prompt}
\fi

\medskip\noindent\emph{System prompt (used when \texttt{instruct=True}):}

\ifverboseprompts
\begin{prompt}{GSM8K -- Select Best Correct Completion (System)}
You are an expert evaluator of mathematical word problem solutions. Given a
word problem and candidate solutions, determine which demonstrates the best
reasoning quality.

Evaluate on:
- **Problem comprehension**: Does the solution correctly identify all
quantities and relationships?
- **Step-by-step correctness**: Are all calculations accurate?
- **Logical clarity**: Is the reasoning easy to follow?
- **Completeness**: Are all sub-steps documented?
- **Directness**: Is the approach efficient without unnecessary repetition?

Provide a brief comparative analysis, then end your response with exactly:
**Judgment: [IDX]**
where IDX is the 0-based index of the best solution.
\end{prompt}
\else
\begin{prompt}{GSM8K -- Select Best Correct Completion (System, condensed)}
Expert evaluator of word-problem solutions. Assess comprehension, step
correctness, logical clarity, completeness, and directness.
End with: **Judgment: [IDX]**
\end{prompt}
\fi

\medskip\noindent\emph{User prompt (used when \texttt{instruct=True}):}

\ifverboseprompts
\begin{prompt}{GSM8K -- Select Best Correct Completion (User)}
**Problem:** {problem}

**Solutions (indices 0-{max_idx}):**
{solutions}
\end{prompt}
\else
\begin{prompt}{GSM8K -- Select Best Correct Completion (User, condensed)}
**Problem:** {problem}   **Solutions (indices 0-{max_idx}):** {solutions}
\end{prompt}
\fi